\documentclass[final,5p,times,twocolumn,authoryear,nopreprintline]{elsarticle}

\usepackage{subfigure}
\usepackage{setspace}
\usepackage{geometry} 
\usepackage{framed,multirow}
\usepackage{latexsym}
\usepackage{bm}
\usepackage{amsmath}
\usepackage{booktabs}
\usepackage{xcolor}
\usepackage{array}
\usepackage[ruled,vlined]{algorithm2e}
\usepackage{dblfloatfix}
\usepackage{verbatim}
\usepackage{mathrsfs}
\usepackage{amsthm,amsmath,amssymb}
\usepackage{textcomp}
\usepackage{varwidth}
\usepackage{enumitem}
\usepackage{lineno,hyperref}
\usepackage{geometry}
\newcommand{\figcaption}{
\setlength{\abovecaptionskip}{0pt}%
\setlength{\belowcaptionskip}{0pt}%
\caption}

\newcommand{\tablecaption}{
\setlength{\abovecaptionskip}{0pt}%
\setlength{\belowcaptionskip}{2pt}%
\caption}

\journal{ISPRS Journal of Photogrammetry and Remote Sensing}

\begin{document}

\begin{frontmatter}

\title{A new weakly supervised approach for ALS point cloud semantic segmentation}

\author[polyu]{Puzuo Wang}
\ead{puzuo.wang@connect.polyu.hk}
\author[polyu,polyusri]{Wei Yao\corref{cor1}}

\ead{wei.hn.yao@polyu.edu.hk}


\address[polyu]{Dept. of Land Surveying and Geo-Informatics,
The Hong Kong Polytechnic University, Hung Hom, Kowloon, Hong Kong SAR, China}
\address[polyusri]{The Hong Kong Polytechnic University Shenzhen Research Institute, Shenzhen, China}

\cortext[cor1]{Corresponding author.}

\vskip 1cm

\begin{abstract}
While there are novel point cloud semantic segmentation schemes that continuously surpass state-of-the-art results, the success of learning an effective model usually rely on the availability of abundant labeled data. However, data annotation is a time-consuming and labor-intensive task, particularly for large-scale airborne laser scanning (ALS) point clouds involving multiple classes in urban areas. Thus, how to attain promising results while largely reducing labeling works become an essential issue. In this study, we propose a deep-learning based weakly supervised framework for semantic segmentation of ALS point clouds, exploiting potential information from unlabeled data subject to incomplete and sparse labels. Entropy regularization is introduced to penalize the class overlap in predictive probability. Additionally, a consistency constraint by minimizing difference between current and ensemble predictions is designed to improve the robustness of predictions. Finally, we propose an online soft pseudo-labeling strategy to create extra supervisory sources in an efficient and nonpaprametric way. Extensive experimental analysis using three benchmark datasets demonstrates that in case of sparse point annotations, our proposed method significantly boosts the classification performance without compromising the computational efficiency. It outperforms current weakly supervised methods and achieves a comparable result against full supervision competitors. For the ISPRS 3D Labeling Vaihingen data, by using only 1\textperthousand{} of labels, our method achieves an overall accuracy of 83.0\% and an average F1 score of 70.0\%, which have increased by 6.9\% and 12.8\% respectively, compared to model trained by sparse label information only.
\end{abstract}


\end{frontmatter}

\section{Introduction}\label{sec:introduction}
As an important data source of active remote sensing, airborne laser scanning (ALS) data depicts a precise three-dimensional representation of large scale out-door scenes. While 3D coordinates and associated attributes (e.g. laser reflectance and return count information) are usually contained in ALS point clouds, to fully interpret a complex geographical scene, the key step is to acquire semantic information as a valuable cue utilized in a variety of remote sensing applications, such as land cover survey~\citep{YAN2015295}, forest monitor~\citep{YAO2012368}, change detection~\citep{OKYAY2019102929} and 3D mapping~\citep{ZHANG201886}. Semantic segmentation, or classification, usually  assigning a label to each point, is an indispensable solution for point cloud parsing  .

Over past decades, point cloud semantic segmentation is always a research hot-spot in scene understanding. Initially, former studies focused on developing rule-based methods to distinguish different categories of land covers~\citep{ANTONARAKIS20082988, 6497495}. Hand-crafted features were extracted based on characteristics of specific point clouds, and the hard threshold was used to classify different objects. While these methods were unsupervised, the performance relied heavily on effective feature extraction and suitable threshold settings, which makes them hard to generalize to new areas. The application of machine learning methods to the classification of point clouds improved the accuracy of results. In classical machine learning methods~\citep{GUO201156,WEINMANN2015286}, well-designed hand-crafted features were still required, but they were fed into supervised classifiers to optimize model parameters automatically. The development of convolutional neural networks (CNNs) has apparently driven the progress of point cloud semantic segmentation tasks, and  cutting-edge methods based on deep learning have been newly developed, achieving state-of-the-art results~\citep{qi2017pointnet, pointnet++, dgcnn, thomas2019kpconv}. Most of them have focused on designing new network structures or convolution kernels based on the characteristics of point cloud data, without taking into account the high costs paid to secure the availability of labels. Actually, the supervised methods rely on a large amount of precise data annotations, which triggers the issue of data hungry~\citep{9430028}. Data hungry refers to the demand on large number of labeled data for supervised learning to achieve leading results. To collect precise annotations is usually associated with heavy workloads, even requiring extremely meticulous efforts for an expert operator to complete the task. Moreover, the labeling of ALS point clouds is particularly difficult, usually demanding the operator to confirm the category of a point from multiple perspectives. The occlusions caused by the scan pattern of ALS systems lead to data voids, which makes it even harder to determine the exact label of points in occluded areas. In addition, the discrete data structure of point clouds in 3D space increases the difficulty of visual interpretation.

\begin{figure}[tb]
\begin{center}
       \subfigure[]{
                \begin{minipage}{.45\linewidth}
                \centering
                \includegraphics[width=1.0\columnwidth]{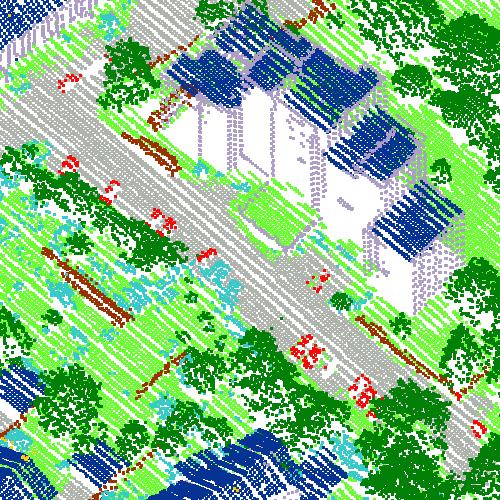}
                \end{minipage}
                }
      \subfigure[]{
                \begin{minipage}{.45\linewidth}
                \centering
                \includegraphics[width=1.0\columnwidth]{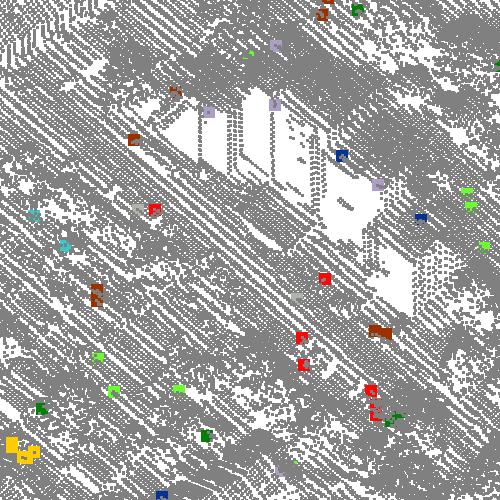}
                \end{minipage}
                } 
      \subfigure[]{
                \begin{minipage}{.45\linewidth}
                \centering
                \includegraphics[width=1.0\columnwidth]{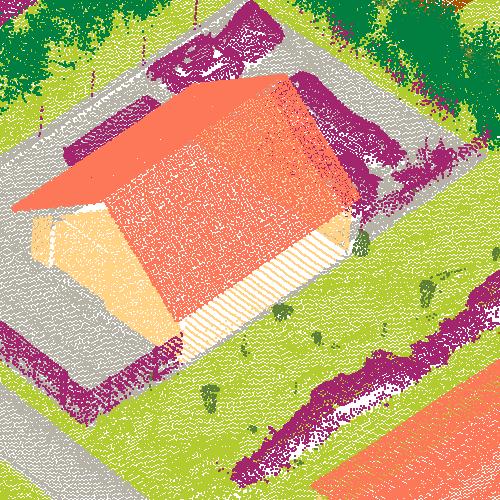}
                \end{minipage}
                }
      \subfigure[]{
                \begin{minipage}{.45\linewidth}
                \centering
                \includegraphics[width=1.0\columnwidth]{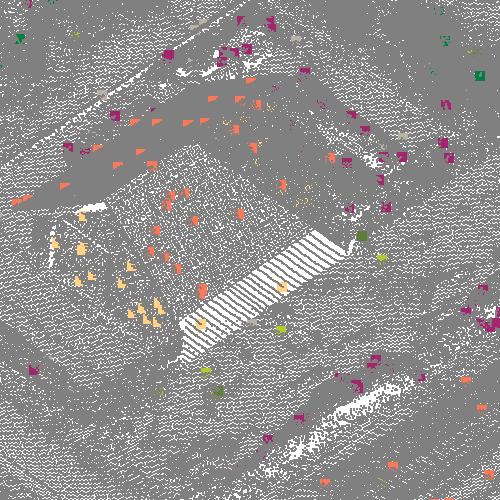}
                \end{minipage}
                }           
	\figcaption{Comparison of fully- and weakly-labeled point clouds. (b) and (d) present the situation in which only 1\textperthousand{} of points are labeled. The size of weak labels in (b) and (d) is enlarged for better visualization.}
\label{fig:weaklabel}
\end{center}
\end{figure}

Although data labeling is a difficult and time-consuming job, the collection cost of massive unlabeled point cloud data is greatly reduced, thanks to the advances in LiDAR technology and diversified data acquisition platforms. So far, multiple ALS point clouds benchmark datasets have been released for the task of semantic segmentation. For instance, a new benchmark, Hessigheim 3D  Benchmark (H3D)~\citep{KOELLE2021H3D}, contains tens of millions of high-resolution 3D point clouds, the density of which is about 800 pts/m². Confronted with massive point clouds subject to annotation, we naturally ask a question, whether promising results can be achieved without the necessity to label the entire scene, and how is the performance of methods affected under such condition? An illustrated comparison of fully labeled and sparsely labeled data is shown in Fig.~\ref{fig:weaklabel}. If competitive classification results could be achieved by only using incomplete labels, the workload of data annotation will be considerably reduced, contributing significantly to the efficiency in real-life applications.

Semi- and weakly supervised learning are commonly used methods addressing situations, in which the scarcity of labels prevails. As two concepts are often mixed up in studies, we use weakly supervised learning or weak supervision in this study to represent the situation in which label information is incomplete and deficient. Recently, comprehensive reviews of weakly supervised learning methods are performed~\citep{zhou2018brief,van2020survey}. Under weak supervision, it can be categorized into two types, transductive and inductive mode. For transductive one, unlabeled data are exactly test one and also used as co-training set in parallel to parameterize the model. In contrast, inductive learning follows the normal process of a full supervision scheme, training a model separately with labeled samples and generalizing to unseen areas. The mode of inductive learning is considered for investigation in our study due to high 
applicability. By applying a weakly supervised method, it is meaningful to choose a suitable weak label annotation strategy for the purpose of reducing labeling workload. In image processing, weak labels are represented as a few labeled images~\citep{dong2018few}, a few labeled pixels~\citep{bearman2016}, labels of image patches~\citep{7452558}, or a few bounding boxes or categories that appear in images~\citep{Kolesnikov2016}. For conventional computer vision tasks, image-level labels are the most accessible annotation, whereby adding a portion of labeled pixels could further contribute to an improved classification result. However, it is different for ALS point cloud processing, which usually covers a large-area earth surface. The data has to be subdivided into small blocks to enable training. Thus, compared with annotating scene-level labels for large number of extracted subsets, directly labeling few points across the whole data seems more practical. Given a fixed budget number of labeled points, how to choose an effective labeling strategy is another issue.~\citet{xu2020weakly} presented theoretical and experimental analysis to demonstrate the superiority of discrete labels against conventional ones. Thus, the strategy to initialize spatially discontinuous weak-labels is adopted in this study. As for weakly supervised learning methods, while image processing witness a large number of relevant works, there are quite few studies related to point cloud processing.~\citet{guinard2017weakly,yao2020pseudo} proposed a transductive learning framework to classify point clouds with sparse weak labels, which was deemed less practicable. A weakly supervised strategy for point cloud semantic segmentation were developed in~\citep{xu2020weakly}. However, at the least 10\% of labels in a scene were needed to achieve a satisfactory classification result in experiments, so the considerable workload of labeling work was still required, especially for datasets with high density. Our previous study~\citep{wang2021} proposed a pseudo-label-assisted approach to point cloud semantic segmentation using limited annotations, but the training process was inefficient. Additionally, the classification result lacks of robustness when only adopting pseudo-labels.

In this study, we propose a plug-and-play weakly supervised framework for ALS point cloud semantic segmentation, which is integrated with diverse deep network architectures to leverage information in unlabeled data. Firstly, considering that there is no ground-truth information for unlabeled points, we take advantages of the entropy of predictions during training, and entropy regularization is adopted to reduce the class overlap in prediction probability and improve prediction confidence. Then, we develop an ensemble prediction constraint to acquire more robust results by making a contrastive pair, the prediction at current training step with its ensemble value during the whole training process. An online soft pseudo-labeling strategy is proposed to create extra supervisory sources and further improve the accuracy of classification result. Pseudo-labels are generated from the ensemble prediction, and their weights contributing to the loss value is based on the reliability calculated from the predicted probability. We use KPConv~\citep{thomas2019kpconv}, a point convolution network, as backbone network. Experiments in three ALS datasets indicate that competitive results are achieved compared to those under the full supervision scheme with only 1\textperthousand{} of labels. Our main contributions are summed up as follows:

\begin{itemize}[leftmargin=*]
\item[$\bullet$] We propose a plug-and-play weakly supervised framework for ALS point cloud semantic segmentation, which can be flexibly integrated with mainstream backbone networks by reducing the reliance on label abundance to achieve a competitive result;

\item[$\bullet$] The entropy regularization is introduced to penalize class overlap in predictive probability caused by scant annotations and improve the confidence of predictions;

\item[$\bullet$] A consistency constraint is designed by minimizing the contrast between current prediction and the ensemble one to increase the prediction robustness;

\item[$\bullet$] An online soft pseudo-labeling strategy is proposed to supply additional supervisory sources, which enables the training process to be completed in an efficient and parameter-free mode.
\end{itemize}

The remainder of this paper is organized as follows: In Section~\ref{sec:relatedwork}, we systematically review deep learning-based methods for ALS point cloud semantic segmentation and semi- and weakly supervised learning for image and point cloud semantic segmentation. Our methodology is described in detail in Section~\ref{sec:3}. Section~\ref{sec:experiment} presents the datasets and weak-label settings. In Section~\ref{sec:result}, we conduct extensive experimental analysis to compare and analyze the effectiveness of proposed method. Concluding remarks are provided for future work in Section~\ref{sec:conclusion}.

\section{Related work}\label{sec:relatedwork}

\subsection{Point cloud semantic segmentation}

No matter of full or weak supervision, it is essential to develop a powerful semantic segmentation network structure to extract representative features. Owing to the irregular data distribution in point clouds, the network structure is not as uniform as standard 2D convolutional neural network (CNN), leading to different types of networks. We summarize three main types, including projection-based, voxel-based, and point-based ones.

\subsubsection{Projection-based methods}
Due to the structural irregularity of point clouds, 2D CNNs cannot be directly utilized for point cloud processing. To achieve it, early studies projected point clouds onto images and applied mature 2D CNNs to the classification task. Classical point cloud features were usually considered, such as color, intensity and height.~\citet{Hu2016} proposed a ALS point cloud filtering algorithm by transforming each point into a image. The color value of each pixel was generated based on three types of height differences in a neighborhood of the center point. In~\citet{yang2017}, hand-crafted features of point clouds were extracted to generate images, which were fed into a 2D CNN to achieve point cloud classification.~\citet{yang2018} achieved a better classification result by extension of developing a multi-scale CNN. Meanwhile, a similar study was proposed in~\citet{zhao2018}. One problem of above studies is that each point needs to be converted into one image, associated with low computational efficiency.~\citet{boulch2018snapnet} solved the problem by generating multi-view images of point clouds and fed them into a 2D semantic segmentation network. Then, the classification of each point was acquired with back-projection. ~\citet{Rizaldy2018FULLYCN} utilized the image generation strategy in~\citet{Hu2016} and implemented a fully convolutional network to perform ALS ground filtering, which significantly reduced the computational cost. 

\subsubsection{Voxel-based methods}
Such methods voxelize point clouds and use 3D CNNs to process voxel data. Compared with the projection-based method, the voxel-based method can inherently retain the 3D structural information of the point cloud. VoxNet~\citep{vox2015} transformed points to 3D voxels and achieved 3D object recognition by implementing a 3D CNN. VoxelNet~\citep{zhou2018voxelnet} proposed a 3D detection network and presented an efficient strategy to process sparse point structure. To reduce memory footprint and computation consumption during voxelization process, several point cloud organization structures were integrated in 3D CNNs, such as KD-trees~\citep{Klokov2017863} and octree~\citep{wang2017cnn}. In addition, voxels and projections were combined in some studies.~\citet{qi2016volumetric} analyzed 3D volumetric CNNs versus multi-view CNNs and proposed two new architectures of volumetric
CNNs for 3D object classification.~\citet{qin2019semantic} combined voxel and pixel representation-based networks to classify ALS data. The disadvantage of projection-/voxel-based methods is that point clouds need to be converted into regularized data formats, which inevitably destroyed original geometric structure. 

\subsubsection{Point-based methods}
Recently, point-based networks have established as mainstream method for point cloud semantic segmentaiton. PointNet and PointNet++~\citep{qi2017pointnet,pointnet++} were pioneers in the development of shared multilayer perceptrons (MLPs) to directly analyze point clouds. Randla-net~\citep{hu2020randla} analyzed different point cloud downsampling methods and proposed an computational efficient network. For ALS, different studies had made improvements based on exploiting the characteristics of the data.~\citet{YOUSEFHUSSIEN2018191} introduced Pointnet into ALS data classification and implemented a multi-scale fully convolutional network.~\citet{LI202026} proposed a dense connected network for ALS data classification. Hand-crafted features were integrated into the network, and an elevation-attention module was designed to further enhance the representation of semantic features. In~\citet{HUANG202062}, a multi-scale network was developed, combined with a manifold-based feature embedding module and a graph-structured optimization method. GraNet~\citep{HUANG20211} proposed a local convolution module and a global attention module to mine local and global dependencies in point clouds. Graph convolution network is another branch of point-based methods, which constructs a graph through relative spatial positions between points for feature extraction and fusion. In~\citet{dgcnn}, a dynamic graph was constructed and an edge convolution was proposed for local feature extraction.~\citet{landrieu2018large} segmented point clouds into clusters and proposed a graph-based network. Inspired from convolution kernels in 2D CNNs,~\citet{thomas2019kpconv} proposed a point convolution network, in which point kernels were designed to learn local geometric information.

\subsection{Weakly supervised methods}  
\subsubsection{Image processing}
We firstly review weakly supervised methods in image processing where a number of pioneering works were developed. Entropy regularization~\citep{entropy04}, or entropy minimization has proved useful to semi-supervised learning. Pseudo-label, assigning annotations to unlabeled data based on the predictions of current model, is a simple and efficient method to improve the performance of the classification model under weak supervision. An early pseudo-label study was presented by~\citet{lee2013pseudo}.~\citet{Iscen_2019_CVPR} developed a soft pseudo-label method, and the weight was calculated from the entropy of predicted probabilities.~\citet{he2021re} reduced the bias of pseudo-labels caused by the longtailed class distribution on real-world semantic segmentation datasets. Meanwhile, some methods considered the consistency constraint by creating a contrastive sample. A simple strategy was to conduct data augmentation and add a loss function to constrain the feature similarity.~\citet{laine2016temporal} utilized exponential moving average value of prediction during training for comparison, while ensemble model parameters were directly considered in~\citet{NIPS2017_68053af2}.~\citet{miyato2018virtual} proposed an adversarial training to generate more targeted comparison samples. Several proven semi-supervision strategies were combined to further improve the accuracy~\citep{mixmatch,fixmatch}.

\subsubsection{Point cloud processing}
Until now, there have been few works that used weakly supervised methods to classify point cloud data.~\citet{wei2020multi} applied a point class activation map to classify point clouds using only scene-level labels. However, we argue that it is not practical for point cloud data, particularly in outdoor scenes that cover a wide region, because the complete data must be divided into numerous small blocks and categories contained in each block must be specified. In comparison, assigning labels to a few number of points within the entire scene is a more desirable approach.~\citet{polewski2015active} used an active learning method to detect standing dead trees from ALS data combined with infrared images.~\citet{lin2020active} proposed an active and incremental learning strategy for ALS data semantic segmentation, and manual annotation was iteratively added for training. Nonetheless, the setting of weak labels is used to annotate all points falling into tiles, and manual intervention is required during training. A weakly supervised point cloud semantic segmentation framework was recently proposed by~\citet{xu2020weakly}, and an approximate result of fully supervised learning was obtained using 10$\%$ of labels. However, the used weak labels were a spatial aggregation of downsampled full scene labels, signifying still a high workload of labeling. ~\citet{guinard2017weakly} utilized the point cloud segmentation method to improve the classification accuracy with very few labels, but the result largely depended on the segmentation accuracy and the classification of the point cloud was limited to the same area where weak labels were initialized, which was in nature of transductive learning.~\citet{yao2020pseudo} introduced a pseudo-labeling method into point cloud semantic segmentation. However, similar to~\citet{guinard2017weakly}, the framework was also a transductive learning scheme, and the performance of the model has not been verified on unseen data. Our previous study~\citep{wang2021} introduced pseudo-label method into a inductive learning framework and designed a adaptive threshold to generate pseudo-labels. In~\citet{hu2021sqn}, a semantic query network was proposed to share sparse weak-label information in spatial domain by interpolating features from neighboring points.

\begin{figure*}[t!]
\centering
\includegraphics[width=1.0\linewidth]{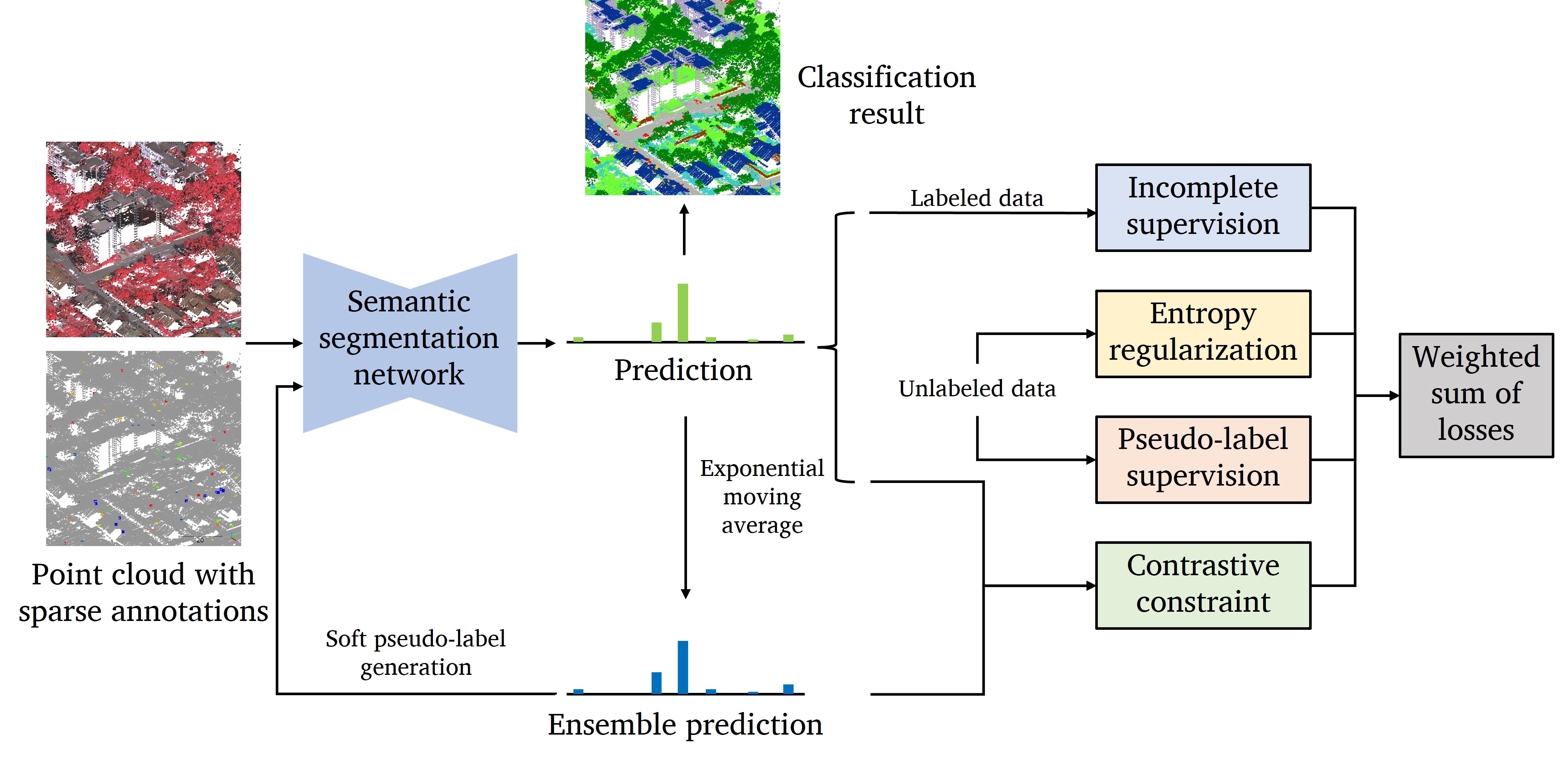}
\figcaption{Schematic diagram of the proposed weakly supervised strategy for ALS point cloud semantic segmentation.}
\label{fig:workflow}
\end{figure*}

\section{Methodology}\label{sec:3}

\subsection{Overview}\label{sec:3.1}
In this study, weak labels are defined as sparse labels randomly distributed across the scene. Given that the input points $P\in\mathbb{R}^{N\times D}$ consist of N points with D dimensional features, M(M$\ll$N) points are assigned labels, denoted as $P_w\in\mathbb{R}^{M\times D}$ and the corresponding labels $L_w\in\mathbb{R}^{M}$ with K classes, and unlabeled points are denoted as $P_u\in\mathbb{R}^{(N-M)\times D}$. In order to compensate for the lack of information caused by sparse annotations, several weakly supervised strategies are proposed to take advantages of both labeled and remaining unlabeled data. The workflow of our method is presented in Fig.~\ref{fig:workflow}. Under weak supervision, the calculation of loss function and backpropogation are implemented only on few labeled points following the scheme of fully supervised learning. Entropy regularization (ER) is adopted to exploit the information of unlabeled data, which can minimize the class overlap and generate predictions with high confidence. Moreover, the limited labels cannot encapsulate the full knowledge of the whole training data, thus often leading to insufficient training process and unstable results. An ensemble prediction constraint (EPC) is developed to enhance the robustness of the trained model by comparing the prediction at current training step with the ensemble value. In addition, the proposed method develops a online soft pseudo-labeling (OSPL) strategy to further improve the performance of the model. Pseudo-labels are generated and updated from ensemble predictions of training set with different weights based on entropy, collaborating with labeled data in calculation of loss function.

KPConv\citep{thomas2019kpconv} is used in this study as backbone network because of its state-of-the-art results achieved on several open datasets. KPConv resorts to the idea of convolution kernels from image processing by extending deformable kernel points to adapt the local features of point clouds. We choose the rigid point convolution kernel and use the same network architecture as KPConv for our semantic segmentation task. The encoder network comprises five convolutional layers, embedding the ResNet-like structure. Skip links are used in the decoder network, and features are passed by the nearest sampling.

\begin{figure*}[t!]
\begin{center}
       \subfigure[]{
                \begin{minipage}{.22\linewidth}
                \centering
                \includegraphics[width=1.1\columnwidth]{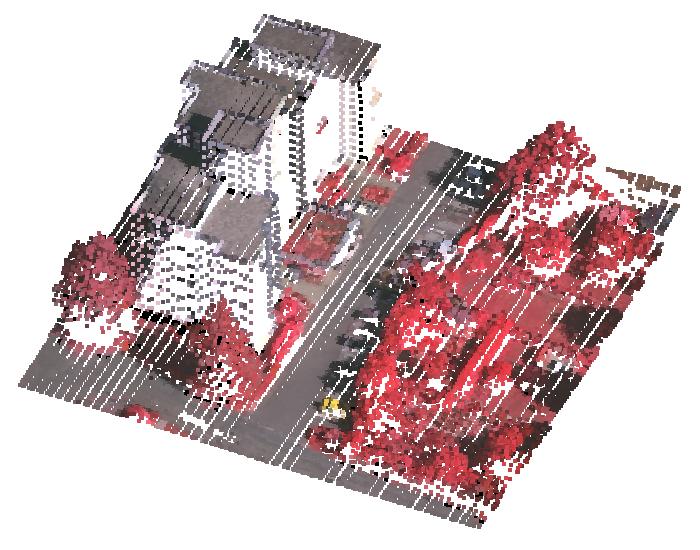}
                \end{minipage}
                }
       \subfigure[]{
                \begin{minipage}{.22\linewidth}
                \centering
                \includegraphics[width=1.1\columnwidth]{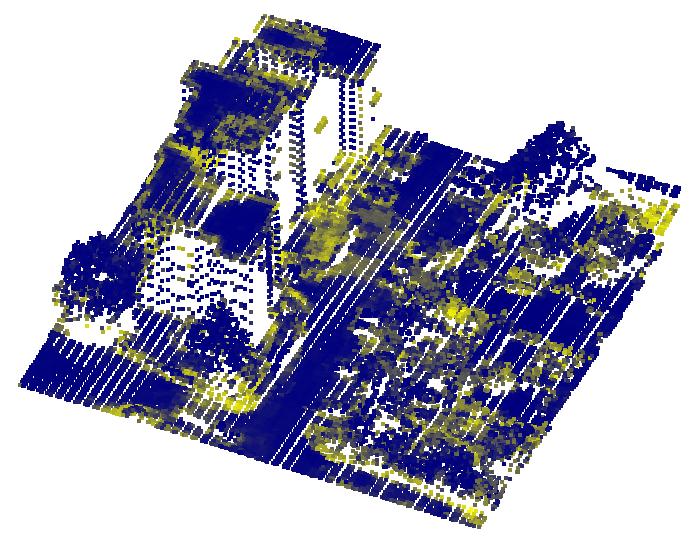}
                \end{minipage}
                }
      \subfigure[]{
                \begin{minipage}{.22\linewidth}
                \centering
                \includegraphics[width=1.1\columnwidth]{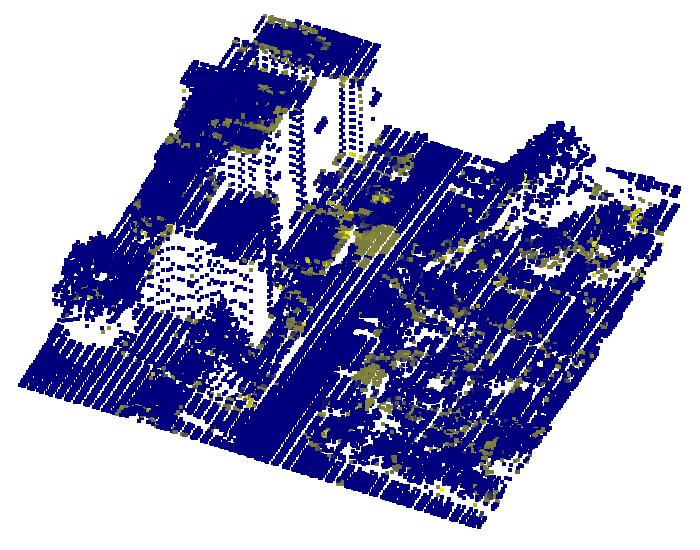}
                \end{minipage}
                } 
      \subfigure[]{
                \begin{minipage}{.242\linewidth}
                \centering
                \includegraphics[width=1.1\columnwidth]{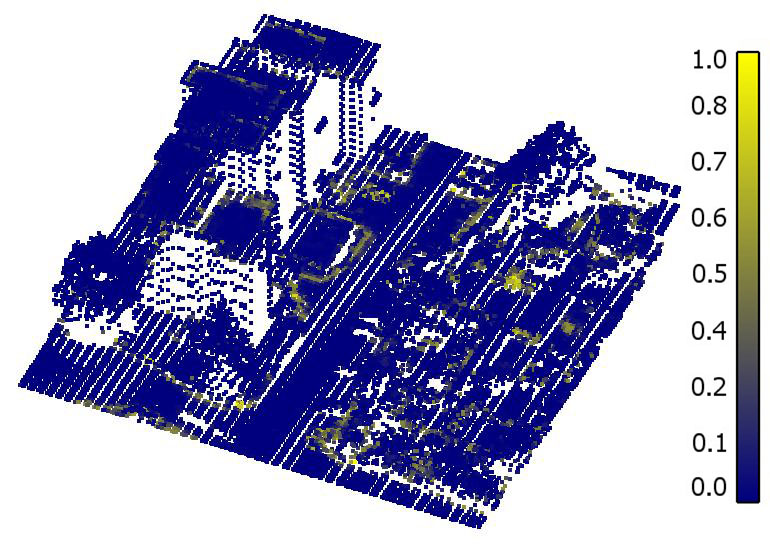}
                \end{minipage}
                }
	\figcaption{Comparison of entropy map of classification result. (a), (b), (c) and (d) present the color map, the result before/after entropy regularization under weak supervision and full supervision, respectively. Low entropy value is colored in blue, while high entropy value is colored in yellow.}
\label{fig:en}
\end{center}
\end{figure*}

\subsection{Entropy regularization}\label{sec:3.2}
In order to present the idea of entropy regularization, we first introduce the incomplete training on labeled data. The training process of incomplete supervision is similar to that of full supervision, and the only difference lies in the design of the loss function. As only a small number of points $P_{w}$ are given label information, we calculate the loss of these points and perform backpropagation. The softmax cross-entropy is a commonly-used loss function in supervised semantic segmentation of point clouds, denoted as:
\begin{equation}\label{equ:1}
	L_{seg} = -\frac{1}{\left | P_{w} \right |}\sum_{i}^{M}\sum_{c}^{K}y_{ic}\log p_{ic}
\end{equation}
where $p_{ic}$ and ${y}_{ic}$ are the prediction and label of point $p_{i}$, respectively.

Notwithstanding lack of annotation information for loss calculation, class-wise posterior probability $p_{c}$ can be predicted for unlabeled points $P_{u}$ by feeding them into the network. Entropy regularization~\citep{entropy04} was proposed based on the conclusion that the information contained in unlabeled data decreases as classes overlap. The entropy $H$ is a measure of class overlap and invariant to the parameterization of the model, which is related to the usefulness of unlabeled data where prediction is ambiguous. Hence, this measure can be utilized to predict well separated classes of unlabeled data. By reducing the class overlap, entropy regularization decreases the uncertainty and supply predictions with a high confidence. Given a target point cloud, the Shannon entropy of each point $H_{p_{i}}$ is used and denoted as:
\begin{equation}\label{equ:2}
	H_{p_{i}}=-\sum_{c}^{K}p_{ic}\log p_{ic}
\end{equation}
The entropy regularization is proposed to minimize entropy of posterior probability, and the loss is calculated as an averaged value:
\begin{equation}\label{equ:3}
	L_{ent} = \frac{1}{\left | P_{u} \right |}\sum_{i}^{N-M} H_{p_{i}}
\end{equation}
Supervised classification loss $L_{seg}$ and entropy regularization loss $L_{ent}$ are respectively calculated on labeled and unlabeled points, delivered to simultaneous optimization during training. A comparison of normalized entropy map $H_{p_{i}}\in [0, 1]^{N}$ of classification result on training set is shown in Fig.~\ref{fig:en}, where a number of points is shown to have a relatively high entropy when initializing the model training with limited weak labels, implying a underfitted model. By contrast, entropy regularization can reduce the entropy for most of points, providing a map similar to that under full supervision.

\subsection{Ensemble prediction constraint}\label{sec:3.3}
In this part, a consistency constraint is proposed to enhance the model for producing more stable predictions. In the case of lack of labels, the consistency constraint, as a self-learning method, refers to minimizing the difference between a pair of well-planned contrastive samples. The motivation is based on the assumption that the prediction $p$ should remain unchanged when applying small perturbations $s$ to the target $t$, such as input data or model parameter, denoted as $p_{t}=p_{t+s}$. One simple and efficient way in point cloud processing is to generate randomly augmented point cloud sample $\hat{P}$ by (rotation, scaling, ...) and feed together with original data $P$ into the network, and assess the distance measure $Dis(P, \hat{P})$ between encoded features or classification probabilities. However, the uncertainty contained in random augmentation leads to unstable predictions. Meanwhile, training such kind of network is usually highly demanding in terms of computational time or resource, since the extra contrastive sample $\hat{P}$ imposes a higher memory burden. To alleviate these issues, several methods borrowed the idea from ensemble learning to create more representative contrastive samples while improving training efficiency. Different from generating contrastive input data, Temporal Ensembling (TE)~\citep{laine2016temporal} recorded the ensemble prediction $\tilde{p}$ during training process and compared it with the prediction $p$ at current training step. As ensemble result is often regarded as a more accurate and robust value, TE provided a better reference target for unlabeled samples for comparison. The shortcoming of TE is that ensemble predictions are updated per epoch, leading to a slow pace of incorporating learned information into the training process. The issue was addressed in Mean Teachers (MT)~\citep{NIPS2017_68053af2} by deriving ensemble model parameters to produce predictions for consistency constraint, in which the ensemble value was updated per step. The trade-off of MT is that two forward propagation calculations at each training step are necessary, reducing the training efficiency. 

\begin{figure}[b]
\centering
\includegraphics[width=1.0\linewidth]{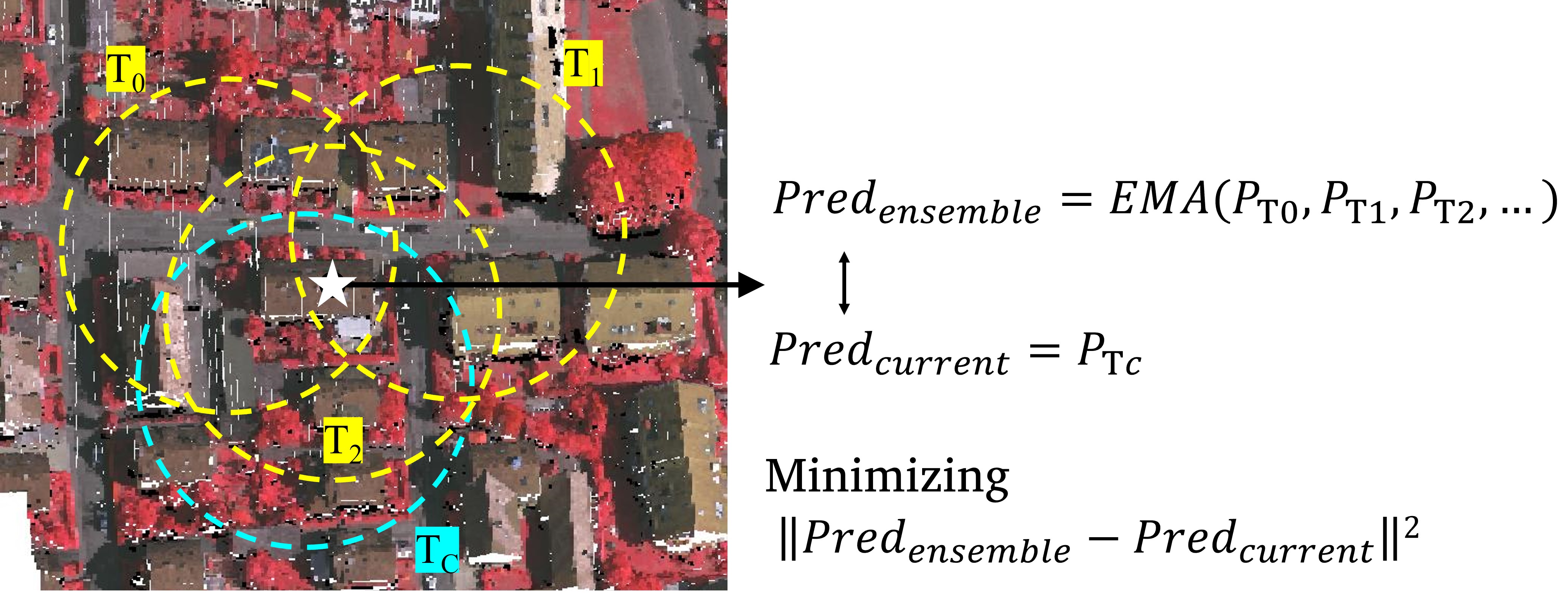}
\figcaption{Illustration of ensemble prediction constraint. $Pred_{ensemble}$ is EMA value of predictions in previous mini-batches, and $Pred_{current}$ is the prediction at the current step. The objective is to minimize the difference between two predictions as the ensemble one is often referred to as more robust estimation.}
\label{fig:epc}
\end{figure}

In this study, similar to TE, ensemble value of predictions $\tilde{p}$ is adopted, and the weakness of TE is alleviated in point cloud classification task using overlapped mini-batches paradigm. Both TE and MT are originally proposed for image processing tasks, whereby typically every image is fed into network only once at each epoch. The situation is different for large-scale point cloud tasks, for which a mini-batch is usually created to cover a small subset of data. Considering inevitable errors at the boundary region of training samples, it requires overlaps between adjacent mini-batches to enhance the feature representation ability. Hence, one point may be fed into the network multiple times at each epoch, contributing to the training reliability and accuracy. As shown in Fig.\ref{fig:epc}, the marked point is assigned to different mini-batches at different training steps. In this way, the ensemble prediction $\tilde{p}_{i}$ is updated multiple times per epoch, and the training efficiency is basically retained as only one forward propagation is needed at each step. Furthermore, points in overlapped training areas covered by multiple mini-batches contain different global information, which can be regarded as a point-wise data augmentation. For each point, exponential moving average (EMA) is utilized for computing ensemble value, which updates the variable related to its historical values. At each step, the ensemble prediction of points $\tilde{p}$ contained in mini-batches will be updated. Given the ensemble prediction of a point for ($t-1$)th training round as $\tilde{p}_{t-1}$, it will be updated in the next round as:
\begin{equation}\label{equ:4}
	 \tilde{p}_{t} = \alpha \cdot \tilde{p}_{t-1} + (1-\alpha)\cdot p
\end{equation}
where $\alpha$ is the coefficient to balance the weighting between ensemble and new values, set to 0.9 in this study. Mean square error (MSE) is utilized to describe the consistency cost, denoted as:
\begin{equation}\label{equ:5}
	 L_{epc}=-\frac{1}{\left | P \right |}\sum_{i}^{N}\left \| p_{i}- \tilde{p}_{i}\right \|^{2}
\end{equation}
where $p_{i}$ and $\tilde{p}_{i}$ are prediction at current step and its ensemble value, respectively. This consistency loss applies to both labeled and unlabeled data.

\subsection{Online soft pseudo-labeling}\label{sec:3.4}
As a simple yet efficient semi-supervised method, pseudo-label $y^{pl}$ can alleviate the problem of limited annotations. The pseudo-label method is a learning form where a classifier is trained to produce predictions, and then retrained by taking inferred classes for unlabeled data as true labels. By increasing the number of pseudo-labels, we intend to reproduce the class distributions in the feature space at scene level. Vanilla pseudo-labeling methods face two problems. Firstly, it appears to inevitably have incorrect predictions in generating and updating pseudo-labels, arising from the underfitted model trained by weakly labeled data. Therefore, a criteria has to be designed to identify reliable predictions and overcome the inaccuracy, which is usually an empirical study considering properties of different networks and data sets. In addition, it needs to iteratively update the pseudo-labels during the entire training process for optimized performance, thus leading to a less efficient training process. In this study, we propose an online soft pseudo-labeling method attempting to solve these two problems, which enables all unlabeled points to be involved in pseudo-label training and maintain quasi the same training speed as the baseline.

\subsubsection{Soft pseudo-label}\label{sec:3.4.1}
Generally, a prediction with high posterior probability is more likely to be correct. Thus, a commonly used method is to select labels with predicted probabilities exceeding a fixed threshold. However, it is difficult to choose a generic threshold applicable to different datasets. In order to better reveal the class distribution and association across whole scene, we derive and soften pseudo-labels on all unlabeled data $P_{u}$ by associating them with different weights $w$ based on the class distribution uncertainty. In this study, the Shannon entropy of predicted probability $H_{p}$ is utilized to measure the uncertainty, and larger value it represents higher uncertainty. We follow the strategy in~\citep{Iscen_2019_CVPR}, and the weight $w_{i}$ for point $p_{i}$ is defined as:
\begin{equation}\label{equ:6}
	 w_{i}=1-\frac{H_{p_{i}}}{\log K}
\end{equation}
where $H_{p_{i}}$ is defined in Equation~\ref{equ:2}, and $\log K$ is used to normalize $w_{i}$ to [0, 1] because it is the maximum value of $H_{p_{i}}$ according to the principle of maximum entropy. 

\begin{figure}[b!]
\begin{center}
       \subfigure[Normal pseudo-labeling]{
                \begin{minipage}{1.0\linewidth}
                \centering
                \includegraphics[width=1.0\columnwidth]{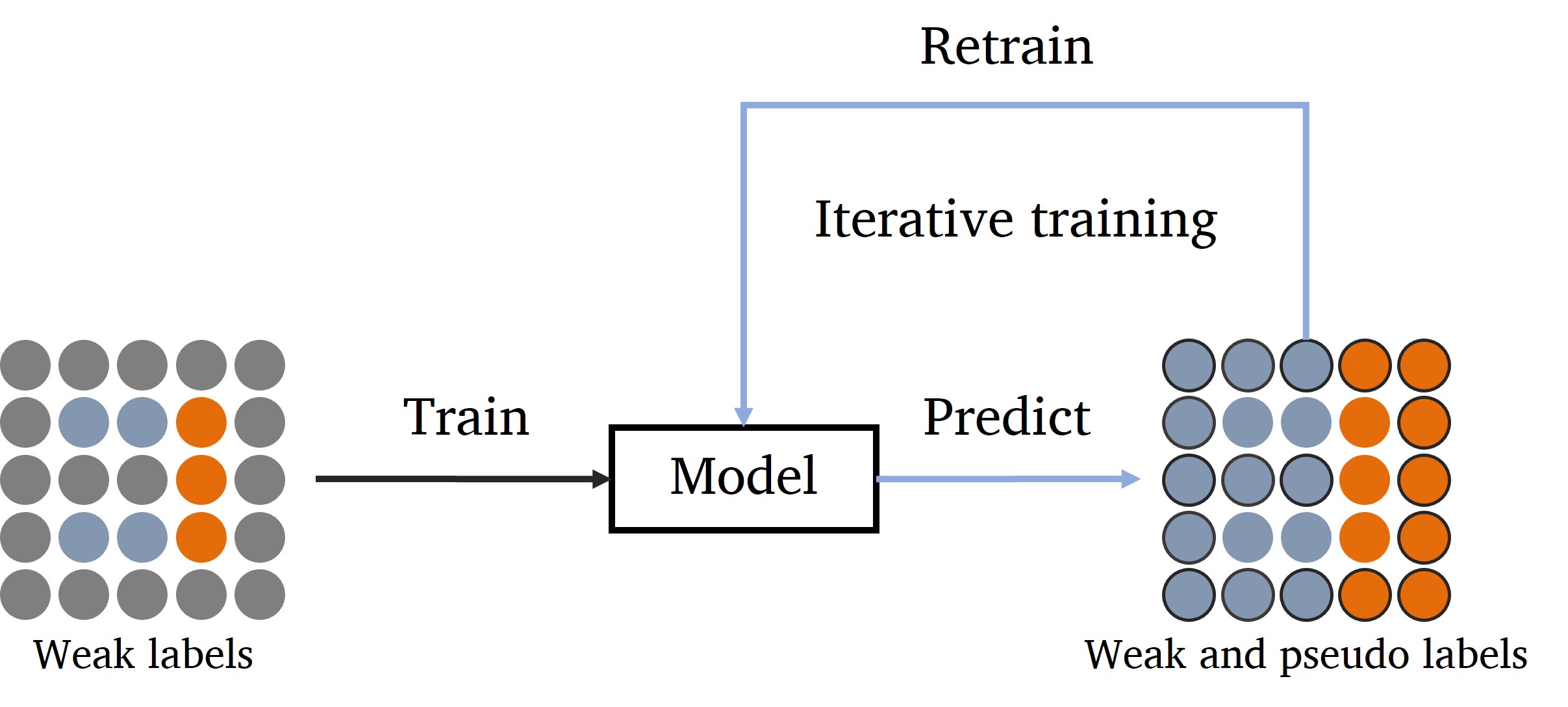}
                \end{minipage}
                }
      \subfigure[Online soft pseudo-labeling]{
                \begin{minipage}{1.0\linewidth}
                \centering
                \includegraphics[width=1.0\columnwidth]{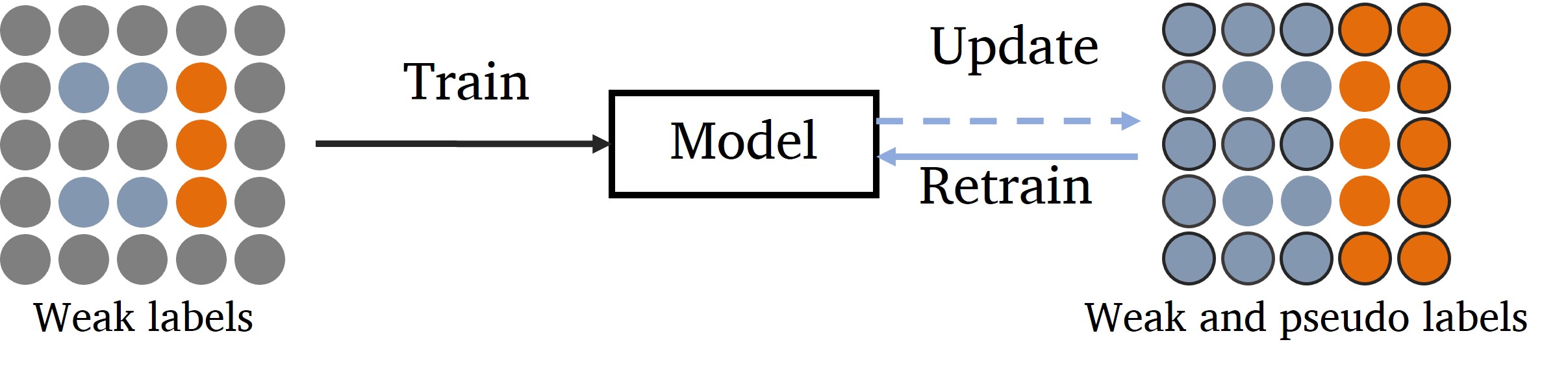}
                \end{minipage}
                } 
	\figcaption{A comparison of training process between normal pseudo-labeling and online soft pseudo-labeling.}
\label{fig:pl}
\end{center}
\end{figure}

\subsubsection{Online training}\label{sec:3.4.2}
Instead of iteratively updating and training, we propose to make use of output predictions from forward propagation to generate pseudo-labels directly. The ensemble prediction during training $\tilde{p}_{i}$ defined in Section~\ref{sec:3.3} is utilized due to its robustness and efficiency. Thus, hard pseudo-labels $y^{pl}$ from unlabeled training data are generated as:
\begin{equation}\label{equ:7}
	 {y_{i}}^{pl} = \arg\max \tilde{p}_{ic}, i\in P_{u}
\end{equation}
We argue that true weak labels from ground truth are essential to model training. Thus, to retain their influence, the loss functions of ground truth and pseudo-labels are formulated separately. The loss of pseudo-labels is calculated by weighted cross entropy using instant predictions, denoted as:
\begin{equation}\label{equ:8}
	 {L_{seg}}^{pl} = -\frac{1}{\left | P_{u} \right |}\sum_{i}^{N-M}w_{i}\sum_{c}^{K}{y_{ic}}^{pl}\log p_{ic}
\end{equation}
A comparison of the training process of two strategies is illustrated in Fig.~\ref{fig:pl}. After acquiring pseudo-labels from predictions on training set, conventional methods need to retrain the model until convergence to update pseudo-labels and iterate the process. In our method, we contemporaneously generate and determine all pseudo-labels in a mini-batch from $\tilde{p_{i}}$ at each step, so its training process is performed quasi equivalent to that of normal supervision, with retention of similar training time.

\paragraph{Relation to ER and EPC}
All three constraints are based on prediction probability during training. Both ER and OSPL can generate high confident predictions, but in a different way. The function of ER is to reduce class overlap in probability distribution, so it particularly benefits unlabeled points with high uncertainty $P_{u}(high\_ent)$. OSPL further extends by generating one-hot labels to reduce the uncertainty of unlabeled data. As soft pseudo-label assigns higher weights to more reliable predictions, OSPL takes advantage of unlabeled points with high confidence and low entropy $P_{u}(low\_ent)$. Hence, ER reduces the number of $P_{u}(high\_ent)$ and produces more $P_{u}(low\_ent)$, and OSPL mainly utilizes $P_{u}(low\_ent)$ to generate reliable pseudo-labels for further improving the trained model. As for EPC, it is utilized to ensure the robustness of predictions by taking ensemble values as reference, forming a smoothness term to constrain ER and OSPL. It is due to the reason that ER and OSPL may also strengthen the impact of incorrect predictions in unlabeled data, and the consistency constraint could offset such confirmation bias.

\begin{algorithm}[b!]
\SetKwData{Left}{left}\SetKwData{This}{this}\SetKwData{Up}{up}
\SetKwFunction{Concatenate}{Concatenate}\SetKwFunction{Kmeans}{K-means}
\SetKwInOut{Input}{Input}\SetKwInOut{Output}{Output}
\Input{Point clouds ${P\in\mathbb{R}^{N\times D}}$, \\ 
        Labels ${y\in\mathbb{Z}^{M}} (M\ll N)$}
\Output{Predictions ${p\in\mathbb{Z}^{N\times C}}$}
\tcp{Stage 1}
\For{$epoch \leftarrow 1$ \KwTo $100$}
{
    \For{each mini-batch $B$}
    {
        Train one step: \\
        $\Theta = \Theta - \eta \nabla (l_{seg}(y,p_{l}|\Theta)
        +\lambda_{ent}l_{ent}(p_{u}|\Theta)
        +\lambda_{epc}l_{epc}(\tilde{p}, p|\Theta))$;\\
	    \# $\Theta$ is learned parameters, $p_{l}$ and $p_{u}$ are predictions in labeled and unlabeled data\\
	    Update ensemble prediction:\\
	    $\tilde{p}_{i}=\alpha \cdot \tilde{p}_{{i}}+(1-\alpha) \cdot p_{i}, i\in B$
    }
}
\tcp{Stage 2}
\For{$epoch \leftarrow 1$ \KwTo $100$}
{
    \For{each mini-batch $B$}
    {
        Generate pseudo-labels:\\
        ${y_{i}}^{pl} = \arg\max\tilde{p}_{i}, i\in B\cap P_{u}$;\\
        Train one step: \\
        $\Theta = \Theta - \eta \nabla (l_{seg}(y,p_{l}|\Theta)
        +l_{ent}(p_{u}|\Theta)
        +l_{epc}(\tilde{p}, p|\Theta))
        +{l_{seg}}^{pl}(y^{pl}, p_{u}|\Theta)$;\\
	    Update ensemble prediction:\\
	    $\tilde{p}_{i}=\alpha \cdot \tilde{p}_{{i}}+(1-\alpha) \cdot p_{i}, i\in B$
    }
}
\caption{Weakly Supervised ALS Data Semantic Segmentation}\label{alg:WeakSupSeg}
\end{algorithm}

\subsection{Training Process}
All losses proposed in previous sections participate in backpropagation simultaneously with different weights. The combined optimization problem is presented as:
\begin{equation}\label{equ:9}
	 \arg \min_{\Theta} L_{seg} + \lambda_{ent}L_{ent} + \lambda_{epc}L_{epc} + \lambda_{pl}{L_{seg}}^{pl}
\end{equation}
where $\Theta$ represents model parameters, and $\lambda_{ent}$, $\lambda_{epc}$, $\lambda_{pl}$ are weighting factors. A ramp-up weight for unsupervised loss components is used in this study. In detail, $\lambda_{ent}$ and $\lambda_{epc}$ are set equal in our experiments, defined with a Gaussian curve $\exp [-5(1-T)^2]$ following~\citep{laine2016temporal}, where T increases linearly from 0 to 1 during ramp-up period, while $\lambda_{pl}$ is set zero during ramp-up period. So, the entire training process consisted of two stages, and each of stage lasts for 100 epochs in our experiments. Stage 1 is the ramp-up period, and $\lambda_{ent}$, $\lambda_{epc}$, $\lambda_{pl}$ are all set 1 during stage 2. The training process is detailed in Algorithm~\ref{alg:WeakSupSeg}.

\section{Experiment}\label{sec:experiment}

\subsection{Dataset and preprocessing}
Three ALS point cloud datasets are chosen in this study for evaluation and analysis, including the ISPRS Vaihingen 3D Semantic Labeling benchmark (ISPRS)~\citep{cramer2010,isprs-annals-I-3-293-2012}, the Large-scale ALS data for Semantic labeling in Dense Urban areas (LASDU)~\citep{ijgi9070450,li2013heihe}, and the Hessigheim 3D Benchmark (H3D) ~\citep{KOELLE2021H3D}. 

\begin{figure*}[t!]
\begin{center}
       \subfigure[]{
                \begin{minipage}{.48\linewidth}
                \centering
                \includegraphics[width=1.0\columnwidth]{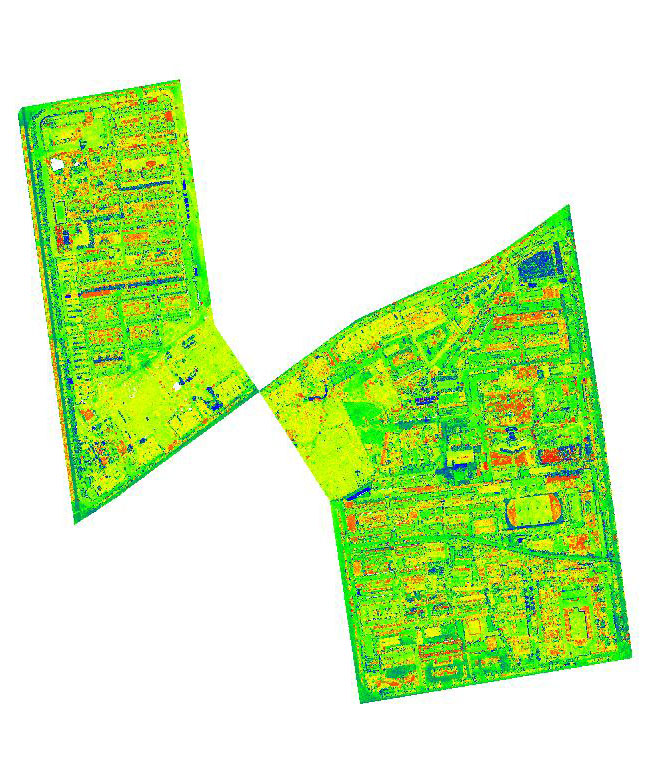}
                \end{minipage}
                }
      \subfigure[]{
                \begin{minipage}{.48\linewidth}
                \centering
                \includegraphics[width=1.0\columnwidth]{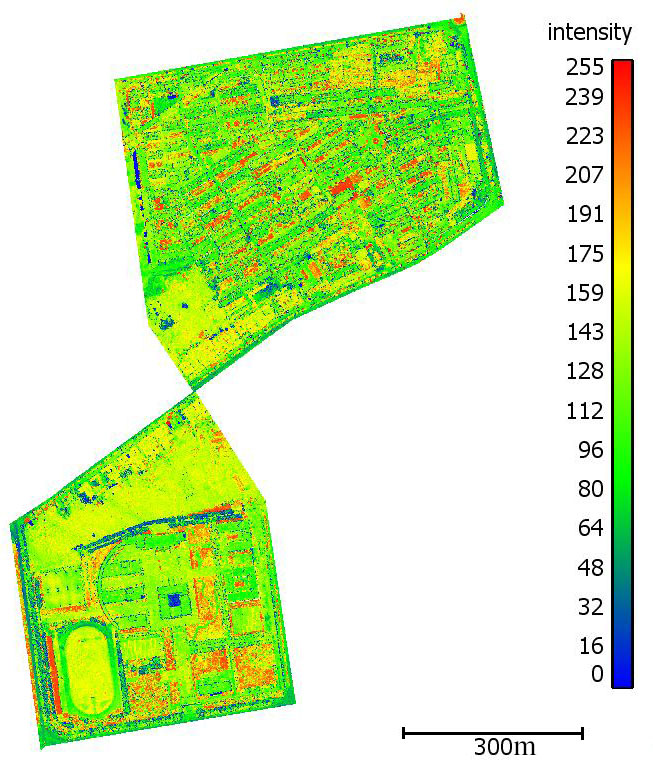}
                \end{minipage}
                } 
	\figcaption{The LiDAR intensity map of training (a) and testing (b) sets of LASDU dataset.}
\label{fig:lasdu}

\includegraphics[width=0.8\linewidth]{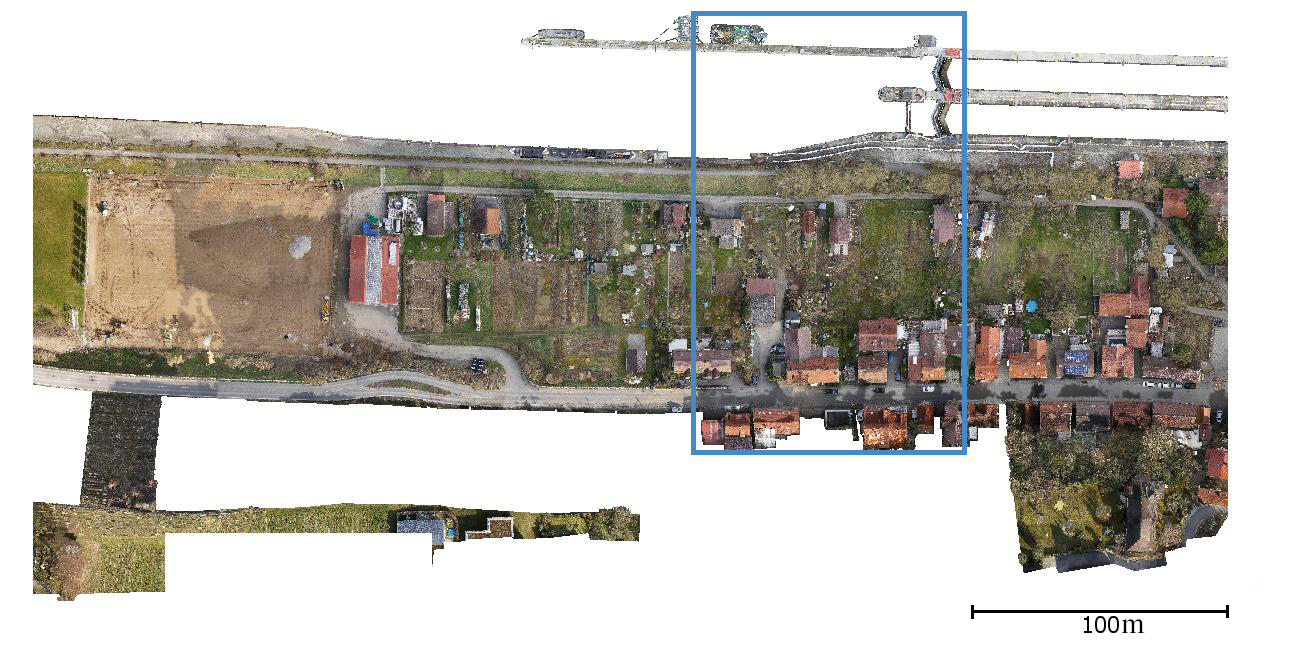}
\figcaption{The RGB color map of training and validation sets of H3D dataset. The validation set is contained in the blue box.}
\label{fig:h3d_color}
\end{center}
\end{figure*}

\paragraph{ISPRS dataset} 
The dataset contains ALS point clouds obtained from the Leica ALS50 system and the co-aligned infrared camera for extracting auxiliary color information. ALS data and aerial images are obtained from Stuttgart region of Germany. The point density of the data is between 4 and 7 points/m². The image data covers the entire area, with a ground sampling distance of 8 cm. There are 9 categories in the dataset, including powerline, low vegetation, impervious surfaces, car, fence, roof, facade, shrub, and tree. The dataset is divided into two parts for training and testing, and the number of points for the training and testing sets is 753,859 and 411,721, respectively. The number of points varies considerably for different classes and are mainly concentrated in the following four categories: low vegetation, impervious surfaces, roof, and tree.

The dataset contains multiple scan data, and the point spacing in the overlap area is very small. To remove redundant points in overlapping ALS strips and maintain an even point density, we set the subsampling grid size to d = 0.4 m and assign labels to deleted points according to the nearest neighbor point in training and testing. The format of utilized  features is \{X, Y, Z, Intensity, IR, R, G\}. 

\paragraph{LASDU dataset}
The dataset is obtained from Leica ALS70 system onboard an aircraft, and the study area is located in the valley along the Heihe River in the northwest of China, which covers an urban area of around 1 km². The average point density is approximately 3–4 points per/m². The whole area is divided into four connected sections, two of which are used as training set, and the remaining two as test set. The number of points is approximately 3.12 million, whereby the training and test parts contain around 0.59 million, 1.13 million, 0.77 million, and 0.62 million, respectively. Five categories are predefined in the dataset, including ground, artifacts, low vegetation, trees and buildings. Fig.~\ref{fig:lasdu} shows the training and testing sets of LASDU dataset. 

Considering the even distribution and relatively low density of point clouds of the LASDU dataset, raw data is directly utilized for training and test. The format of utilized features is \{X, Y, Z, Intensity\}. 

\paragraph{H3D dataset}
The dataset comprises high density LiDAR point cloud of around 800 points/m² enriched by RGB image with a GSD of 2-3 cm, acquired from a Riegl VUX-1LR Scanner and two oblique-looking Sony Alpha 6000 cameras mounted on a RIEGL Ricopter platform. The area of interest is a village of Hessigheim, Germany. The entire area is divided into three connected sections for training, validation and test, respectively. The training and validation sets are used in this study, for which the number of points are approximately 59.4 million and 14.5 million, respectively. Eleven semantic categories are predefined, including low vegetation, impervious surface, vehicle, urban furniture, roof, facade, shrub, tree, soil/gravel, vertical surface, and chimney. Fig.~\ref{fig:h3d_color} shows the color map of training and validation sets of H3D dataset.

Considering the high density of raw data, we set the subsampling grid size to 0.1 m for training and test to improve computational efficiency while preserving point cloud structural details. In inference process, predictions of raw test data are from nearest neighbor interpolation. The format of utilized features is \{X, Y, Z, R, G, B\}.

\begin{figure*}[b]
\begin{center}
       \subfigure[]{
                \begin{minipage}{.48\linewidth}
                \centering
                \includegraphics[width=1.0\columnwidth]{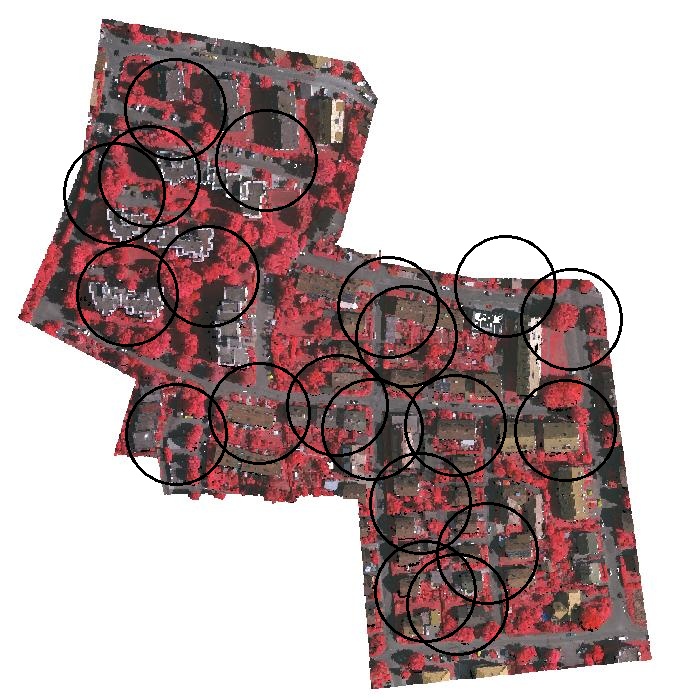}
                \end{minipage}
                }
      \subfigure[]{
                \begin{minipage}{.48\linewidth}
                \centering
                \includegraphics[width=1.0\columnwidth]{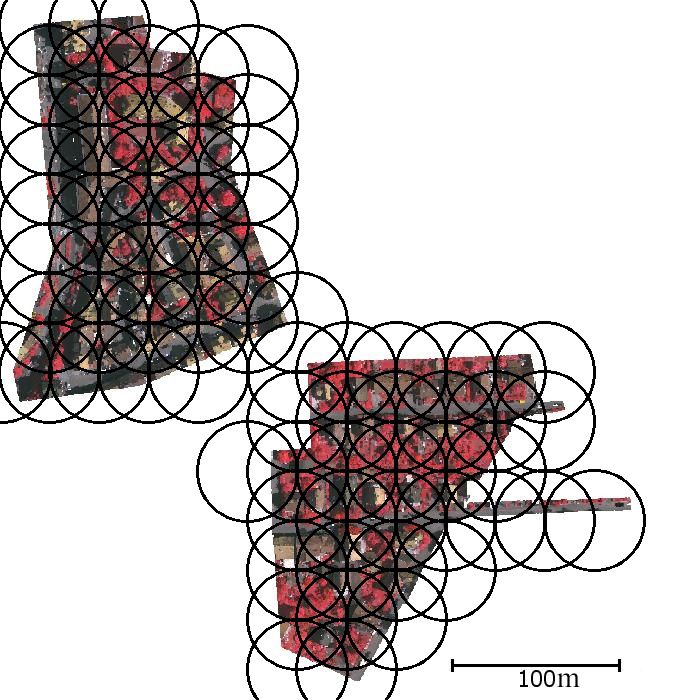}
                \end{minipage}
                } 
	\figcaption{Illustration of data mini-batch generation of the ISPRS dataset, colorized by NIR-R-G spectral. (a) and (b) are training set and test set, respectively.}
\label{fig:block}
\end{center}
\end{figure*}

\subsection{Configuration for weak labels}
To evaluate the performance of our weakly supervised method, weak labels are selected under conditions of different ratios. We randomly initialize point labels for each category with (quasi) the same number, which alleviates the issue of strong class imbalance. In addition, as the exact number of each category remains unknown before data annotation, the proposed weak-label configuration is more applicable to real-world tasks, and no prior knowledge of the probability of class occurrence is contained in weak labels. It should be noted that the number of weak labels for each category is not exactly the same due to the presence of extreme class imbalance in ALS data. Even when the total number of weak labels is quite small, a large proportion of points may be already labeled for certain classes, which is hard to be considered weakly supervised task. We follow three weak-label selection criteria in our experiments:

\begin{itemize}[leftmargin=*]
\item[$\bullet$] Weak labels are randomly selected from the training data;
\item[$\bullet$] The number of weak labels in each category can not exceed $10\%$ of the number of that category;
\item[$\bullet$] Less weak-label cases are contained as subset of the more weak-label cases. For instance, the selected labels in 1\textperthousand{} weak-label setting are fully included in the 2\textperthousand{} weak-label setting.
\end{itemize}
Different weak-label ratios are considered in our experiments, presented in Table~\ref{tab:wl}. We first set a maximum number of weak labels for each of single categories. Then, weak labels are initialized by designed criteria. To approach the exact settings as far as possible, the maximum number can be adjusted. Weak-label initialization for ISPRS and H3D datasets is conducted in subsampled data.

\begin{table}[htb]
    \centering
    \footnotesize
    \setlength\tabcolsep{2pt}%
    \tablecaption{Weak-label settings for data sets}
    \label{tab:wl}
    \subtable[ISPRS dataset]{
        \centering
        \begin{tabular}{cccccc}
        \hline
        Ratio                & 100\%  & 5\textperthousand{}    & 2\textperthousand{}    & 1\textperthousand{}    &5\textpertenthousand{}   \\ \hline
        Max number per class & 98404  & 245   & 95    & 46    & 22    \\
        Sum                  & 401892 & 1992  & 792   & 400   & 198   \\
        Exact ratio          & 100\%  & 4.96\textperthousand{} & 1.97\textperthousand{} & 0.99\textperthousand{}  &4.93\textpertenthousand{}  \\ \hline
       \end{tabular}
       \label{tab:wl_isprs}
       }
    
    \subtable[LASDU dataset]{
        \centering
        \begin{tabular}{cccccc}
        \hline
        Ratio                & 100\%   & 5\textperthousand{}    & 2\textperthousand{}    & 1\textperthousand{}    &5\textpertenthousand{} \\ \hline
        Max number per class & 704425  & 1683  & 677   & 338   & 169   \\
        Sum                  & 1694912 & 8415  & 3385  & 1690  & 845   \\
        Exact ratio          & 100\%   & 4.96\textperthousand{} & 2.00\textperthousand{} & 1.00\textperthousand{} &5.00\textpertenthousand{} \\ \hline
    \end{tabular}
    \label{tab:wl_lasdu}
    }

    \subtable[H3D dataset]{
        \centering
        \begin{tabular}{cccccc}
        \hline
        Ratio                & 100\%    & 2\textperthousand{}   & 1\textperthousand{}   & 5\textpertenthousand{} & 1\textpertenthousand{} \\ \hline
        Max number per class & 4352765  & 2400  & 1200  & 580   & 118   \\
        Sum                  & 12973035 & 24664 & 12664 & 6380  & 1298  \\
        Exact ratio          & 100\%    & 1.90\textperthousand{} & 0.98\textperthousand{} &4.92\textpertenthousand{}  & 1.00\textpertenthousand{}  \\ \hline
     \end{tabular}
     \label{tab:wl_h3d}
     }
\end{table}

\begin{figure*}[b!]
\begin{center}
\includegraphics[width=0.85\linewidth]{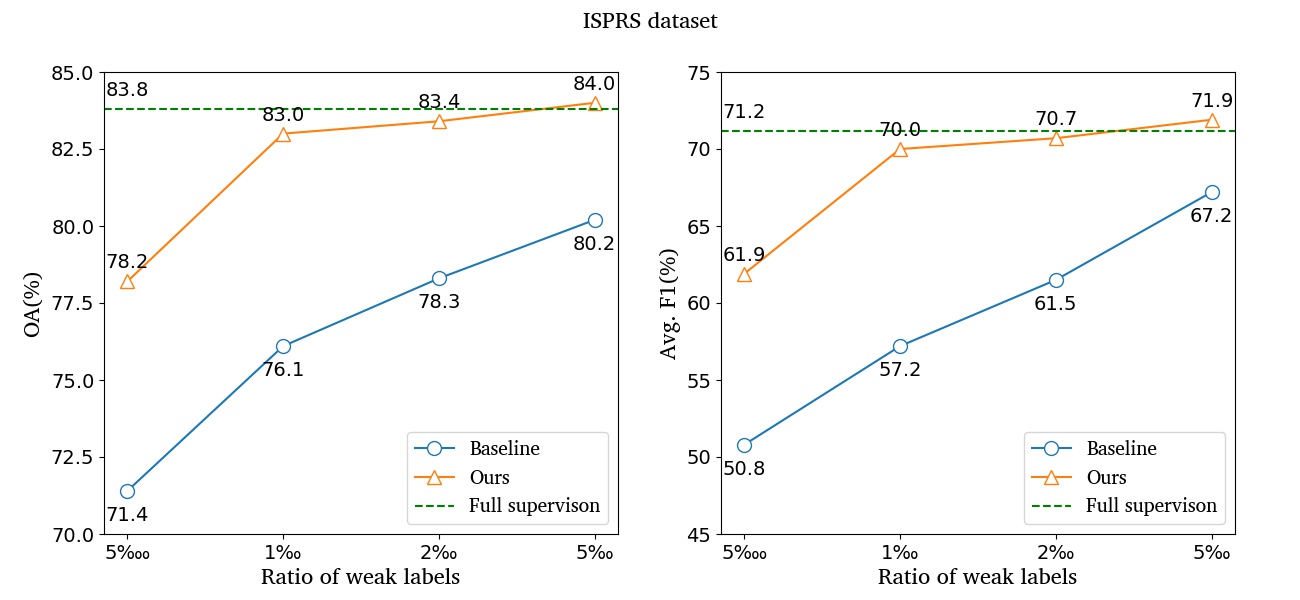}
\figcaption{The impact of the ratio of labeled points on OA and Avg. F1 on ISPRS dataset.}
\label{fig:isprs_line}
\end{center}
\end{figure*}

\subsection{Implementation details}
We follow the strategy in KPConv to generate mini-batch of point clouds for training. An illustration of mini-batch generation is presented in Fig.~\ref{fig:block}. One batch of training sample is defined as a subset of point clouds contained in a circular area, whose radius is related to the point cloud density. As the number of points in each batch is different, the batch size is not fixed, and a upper bound of the sum number of points at each training step is set according to the memory limitation of the graphics card. In our experiments, the radius and number limit for ISPRS and LASDU dataset are the same, 30 m and 120,000, respectively, while those for H3D dataset are 5 m and 90,000. At each epoch, we train 80 steps for ISPRS dataset, and the number for LASDU and H3D dataset are 200 and 400, respectively. During training, different from random selection or uniform block, the location of circle at each batch is determined by a statistical function. Before the training process started, an initial random number is given to each point in training sets, referred to as potential value. Then, the point with minimum potential value is selected as the central position of next batch, and potential value of points in the batch is added by a number between zero to one according to the distance to the center. In this way, the number of times that each point is fed into the model is controlled in a close range, while it increases variety of training samples. An example of mini-batch generation during training is shown in Fig.~\ref{fig:block}(a). We further augment the batch data by randomly rotating points around the z-axis, scaling and adding noise offset in coordinates. During test process, though we have found that the batch generation strategy of training process can benefit the accuracy by increasing the number of mini-batches, to improve the efficiency and maintain a fair comparison with other methods, we create uniform blocks to generate subsets of test data. In detail, circle is still utilized with the radius as settings in training, impelling 50\% overlap between adjacent batches, as shown in Fig.~\ref{fig:block}(b). Then, each point is tested for approximately three times. We use the default parameters of the KPConv segmentation network, adopting Stochastic gradient Descent optimizer, with a momentum of 0.98 and an initial learning rate of $10^{-2}$. All models are implemented in the framework of PyTorch and trained on a single GeForce GTX 1080Ti 11 GB GPU. 

\subsection{Evaluation metrics}\label{sec:Evaluation}

We use the overall accuracy (OA) and F1 score to evaluate the performance of our method. OA is the percentage of predictions correctly classified where the F1 score is the harmonic mean of the precision and recall, presented as:
\begin{equation}\label{equ:10}
\begin{aligned}
& precision = \frac{{tp}}{{tp+ fp}},\\
& recall = \frac{{tp}}{{tp + fn}},\\
& F1 = 2 \times \frac{{precision \times recall}}{{precision + recall}},\\
\end{aligned}
\end{equation}
where $tp$, $fp$, and $fn$ are true positives, false positives, and false negatives, respectively.

\begin{figure*}[t!]
\begin{center}
       \subfigure[]{
                \begin{minipage}{.48\linewidth}
                \centering
                \includegraphics[width=1.0\columnwidth]{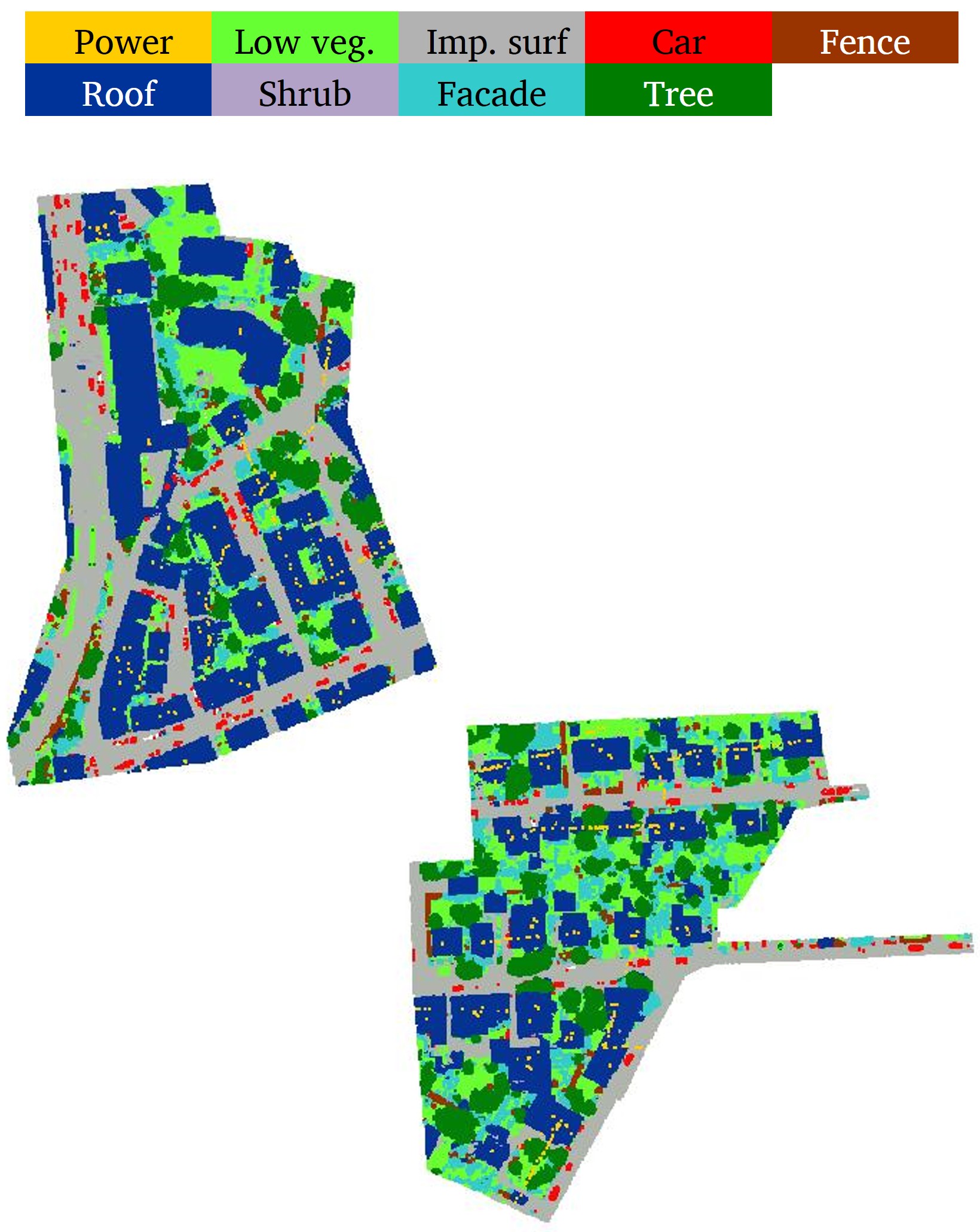}
                \end{minipage}
                }
      \subfigure[]{
                \begin{minipage}{.48\linewidth}
                \centering
                \includegraphics[width=1.0\columnwidth]{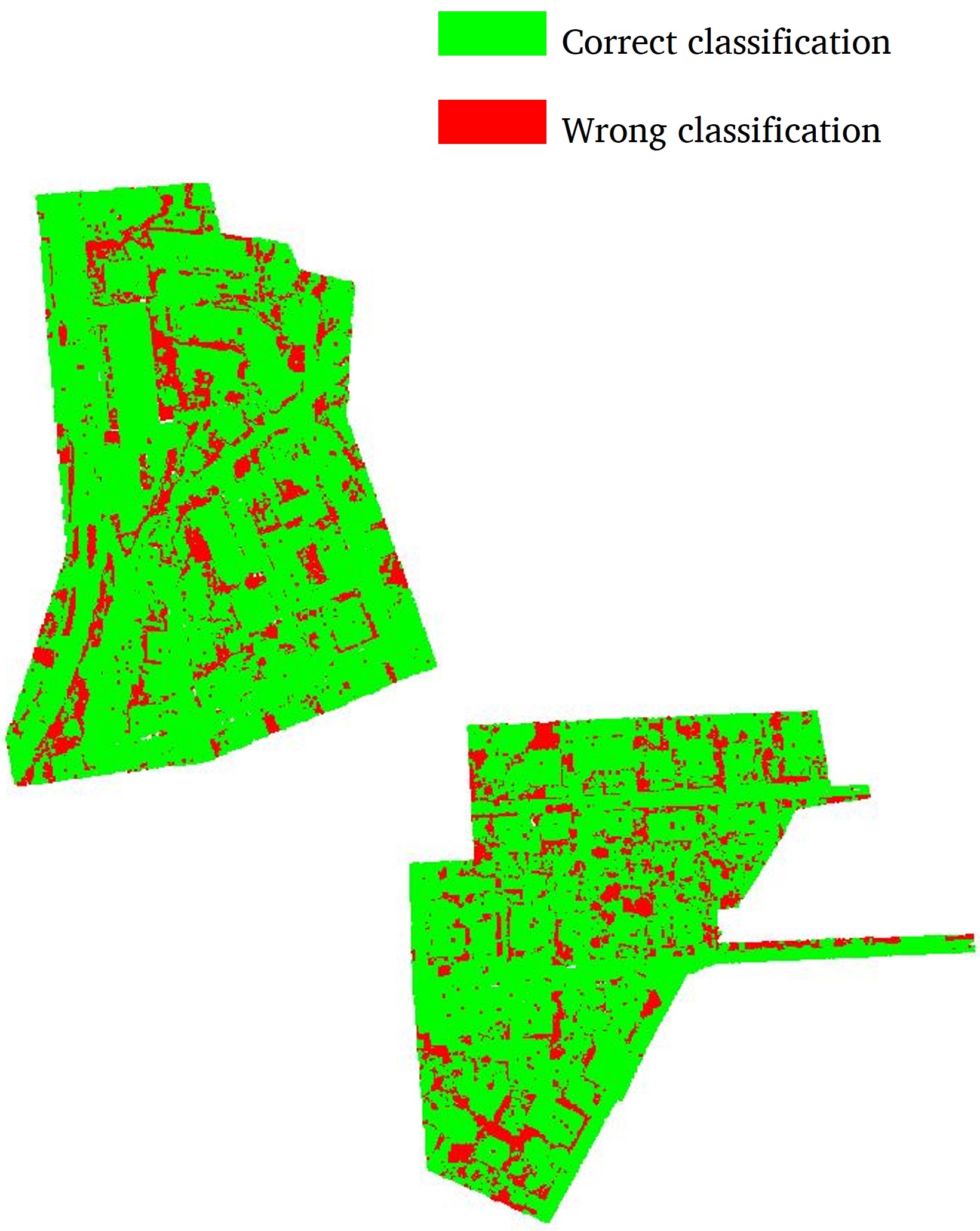}
                \end{minipage}
                } 
	\figcaption{(a) and (b) depict the classification result and the error map of the proposed method at 1\textperthousand{} weak-label on ISPRS dataset, respectively.}
\label{fig:isprs_result&error}
\end{center}
\end{figure*}

\begin{figure*}[t]
\begin{center}
      \subfigure[]{
                \begin{minipage}{.3\linewidth}
                \centering
                \includegraphics[width=1.0\columnwidth]{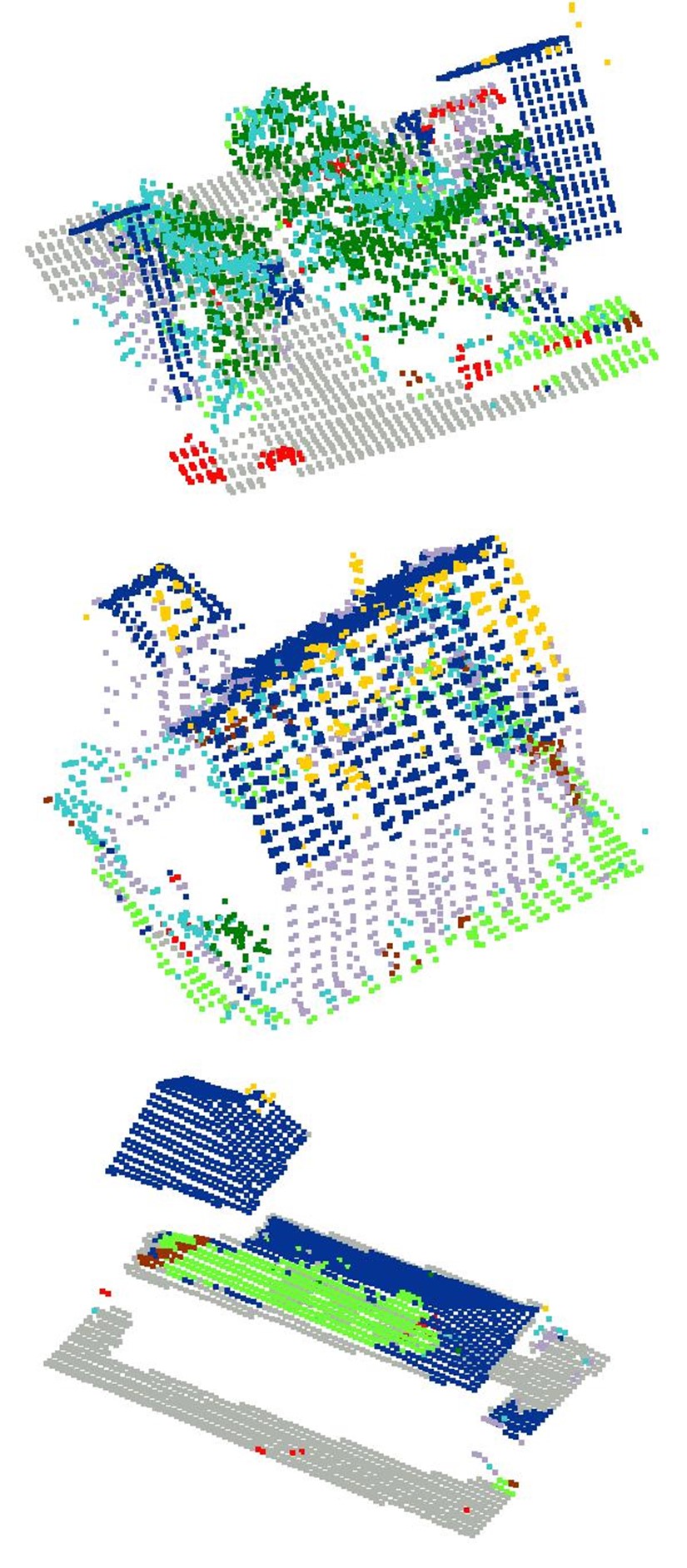}
                \end{minipage}
                }
      \subfigure[]{
                \begin{minipage}{.3\linewidth}
                \centering
                \includegraphics[width=1.0\columnwidth]{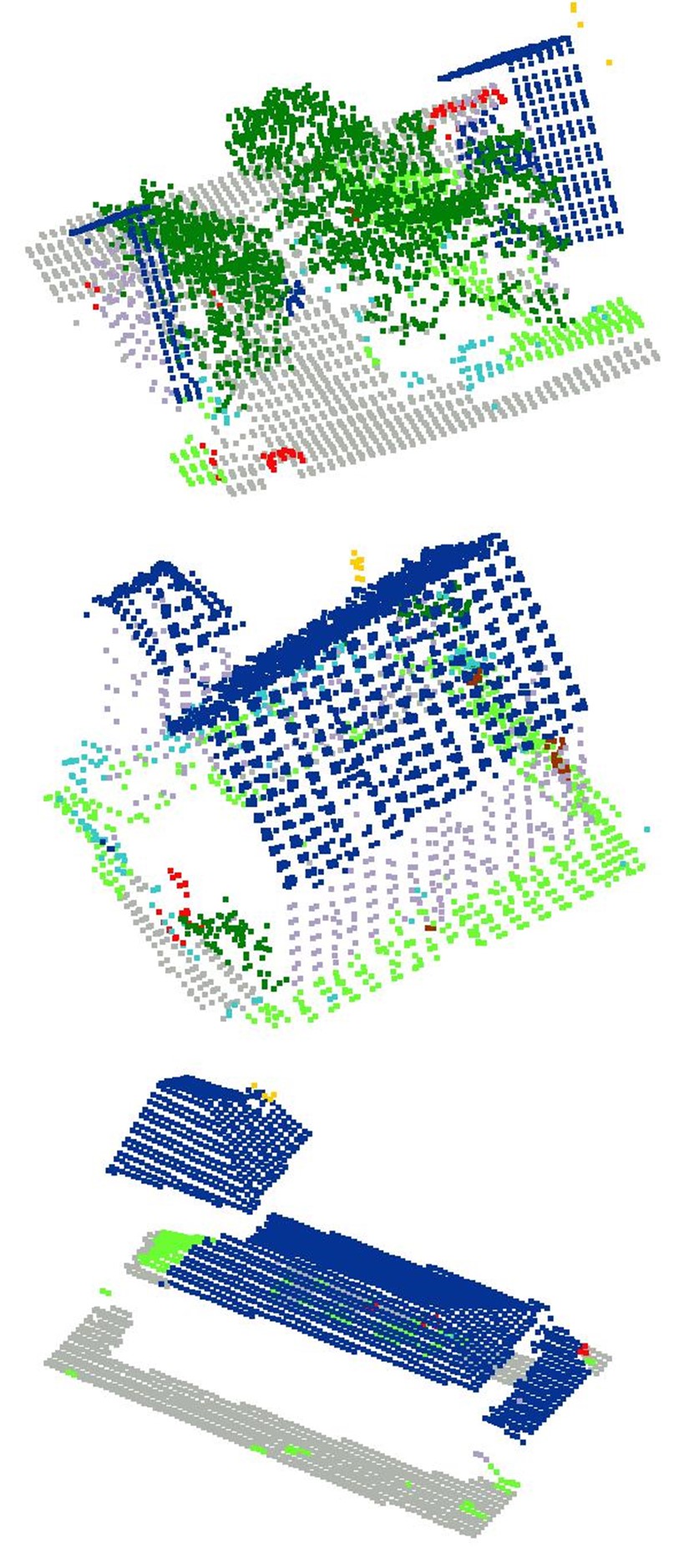}
                \end{minipage}
                } 
    \subfigure[]{
                \begin{minipage}{.3\linewidth}
                \centering
                \includegraphics[width=1.0\columnwidth]{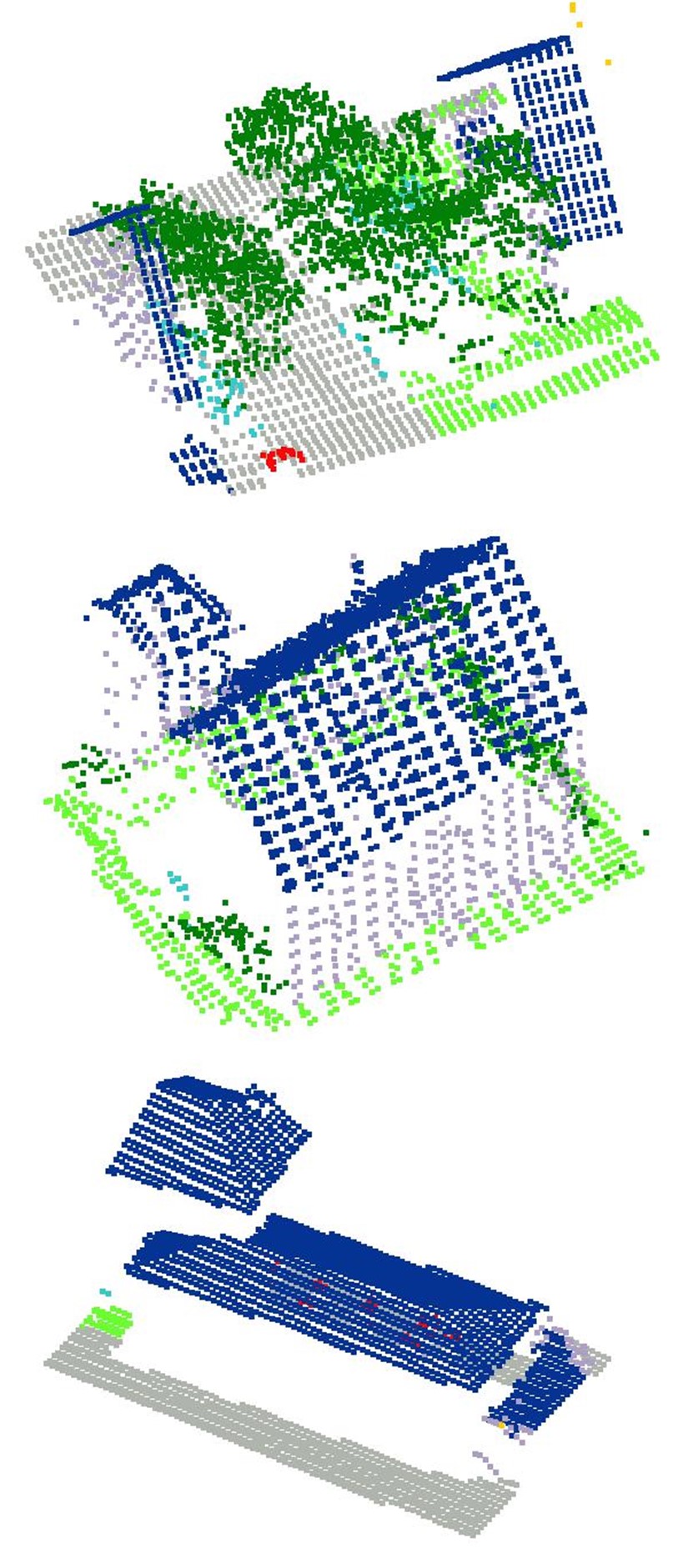}
                \end{minipage}
                } 
	\figcaption{Details of classification result on ISPRS dataset. (a) and (b) represent the results of baseline and our method, and (c) is the ground truth. }
\label{fig:isprs_local}
\end{center}
\end{figure*}

\section{Results and discussion}\label{sec:result}

\subsection{Classification results}
\subsubsection{ISPRS dataset}

\begin{table*}[t!]
\tablecaption{Comparison of full and weak supervision schemes on ISPRS dataset}
\label{tab:isprs}
\centering
\footnotesize
\setlength\tabcolsep{2pt}%
\begin{tabular}{ccccccccccccc}
\hline
\multirow{2}{*}{Setting}        & \multirow{2}{*}{Method} & \multicolumn{9}{c}{F1 Score}                                                & \multirow{2}{*}{Avg. F1} & \multirow{2}{*}{OA} \\ \cline{3-11}
                                &                         & Power & Low veg. & Imp. Surf. & Car  & Fence & Roof & Facade & Shrub & Tree &                          &                     \\ \hline
\multirow{6}{*}{Full Sup.}      & NANJ2                   & 62.0  & \textbf{88.8}     & 91.2       & 66.7 & 40.7  & 93.6 & 42.6   & \textbf{55.9}  & 82.6 & 69.3                     & \textbf{85.2}                \\
                                & WhuY4                   & 42.5  & 82.7     & 91.4       & 74.7 & \textbf{53.7}  & 94.3 & 53.1   & 47.9  & \textbf{82.8} & 69.2                     & 84.9                \\
                                & RandLA-Net              & 69.0  & 80.1     & 91.4       & 73.1 & 36.2  & 93.7 & 58.4   & 45.8  & 82.7 & 70.1                     & 83.1                \\
                                & KPConv                  & 74.2  & 81.7     & \textbf{91.8}       & 80.0 & 29.6  & \textbf{94.7} & \textbf{63.3}   & 44.7  & 80.6 & 71.2                     & 83.8                \\
                                & GANet                   & \textbf{75.4}  & 82.0     & 91.6       & 77.8 & 44.2  & 94.4 & 61.5   & 49.6  & 82.6 & 73.2                     & 84.5                \\
                                & GraNet                  & 67.7  & 82.7     & 91.7       & \textbf{80.9} & 51.1  & 94.5 & 62.0   & 49.9  & 82.0 & \textbf{73.6}                     & 84.5                \\ \hline
\multirow{4}{*}{\begin{tabular}[c]{@{}c@{}}Weak Sup.\\ (1\textperthousand{})\end{tabular}} & Xu \& Lee                     & 13.8  & 63.1     & 58.9       & 22.2 & 23.2  & 79.8 & 27.2   & 30.8  & 52.6 & 41.3                     & 60.6                \\
                                & MT                      & 35.3  & 74.3     & \textbf{90.1}       & 47.0 & 28.3  & 89.2 & 45.2   & 37.3  & 74.0 & 57.9                     & 76.3                \\
                                & Baseline                & 28.1  & 74.1     & 89.6       & 51.0 & 28.5  & 89.0 & 46.0   & 36.8  & 72.0 & 57.2                     & 76.1                \\
                                & Ours                    & \textbf{72.8}  & \textbf{79.3}     & 89.8       & \textbf{75.1} & \textbf{31.5}  & \textbf{95.1} & \textbf{61.5}   & \textbf{43.1}  & \textbf{82.0} & \textbf{70.0}                     & \textbf{83.0}                \\ \hline
\end{tabular}
\end{table*}

We first compare the performance of our method under weak-label settings. It is worth noting that here the baseline method refers to KPConv which only imposes $L_{seg}$ on weak labels. A comparison of experimental results on ISPRS dataset is illustrated in Fig.~\ref{fig:isprs_line}. From the figure, we can see a rise in OA and average F1 score for no matter baseline or our method when gradually increasing the number of weak labels, which is in line with the general perception of relation between accuracy and number of annotations. Compared to baseline, two evaluation indices have significantly improved using our method. Both OA and average F1 score witness rapid growth when the proportion of weak labels goes from 5\textpertenthousand{} to 1\textperthousand{}, and the numbers are 83.0\% and 70.0\%,  marginally below those under full supervision. In the 5\textperthousand{} of weak-label setting, our method achieves an even better result than corresponding full supervision scheme. Considering the trade-off between annotation workload and model performance, we believe that it is a very promising result using 1\textperthousand{} of labels, and the classification result and error map is presented in Fig.\ref{fig:isprs_result&error}. From the figure, it shows that the majority of points are classified correctly, and it maintains a relatively smooth boundary between different objects. Misclassified points show the characteristics of concentrated distribution, and mainly belong to vegetation class including low vegetation, shrub and tree, due to the similar geometric and color information. To provide a detailed information that demonstrate the effectiveness of our method, we visually compare classification results in several local regions in Fig.~\ref{fig:isprs_local}. Due to the limited number of initial weak labels, there are some points that should be easily recognized but are misclassified. Some tree points are wrongly classified to facade by baseline from the first row, and they are corrected by our method. The rest of two rows show our method can rectify most of errors at roof category.

We compare our method with both published results under full supervision and open-sourced weakly supervised methods. It should be mentioned that all these methods are based on deep learning networks. We first introduce deep networks under full supervision. The NANJ2~\citep{zhao2018} method generated images from hand-crafted point cloud features and proposed a multi-sacle CNN based classification method. The WhuY4~\citep{yang2018} method also transformed points into images and performed the classification. The RandLA-Net~\citep{hu2020randla} method proposed a fast point cloud semantic segmentation network by randomly sampling points at each network layer, and a local feature aggregation module was designed to extract semantic features. The KPConv method is the backbone network used for baseline in this study, and we also assess its performance under full supervision. The GANet~\citep{LI202026} method proposed a dense connected network, where a geometry-aware convolution and a elevation-attention module were developed to generate the discriminative deep features. The GraNet~\citep{HUANG20211} method proposed a local encoding convolution and a global attention module to combine local and long-range information. Due to currently limited works related to point cloud weakly supervised learning, we choose two open sourced methods for implementation on utilized datasets. The MT~\citep{NIPS2017_68053af2} method introduced a consistency constraint between predictions produced by current model parameters and their EMA values. MT is originally proposed for image processing tasks, and thus integrated with our baseline, KPConv, to conduct experiments. The Xu \& Lee~\citep{xu2020weakly} method proposed several modules under weak supervision, including a siamese branch, an inexact supervision branch and a smooth branch. 

\begin{figure*}[t!]
\centering
\includegraphics[width=0.85\linewidth]{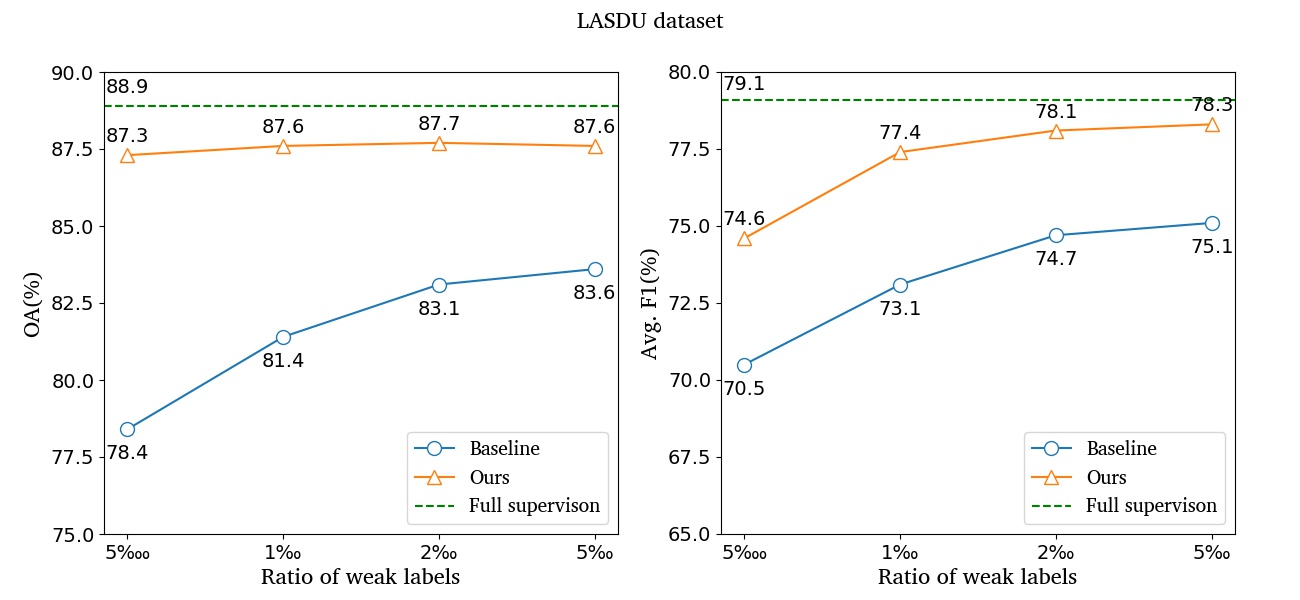}
\figcaption{The impact of the ratio of labeled points on OA and Avg. F1 on LASDU dataset}
\label{fig:lasdu_line}
\end{figure*}

\begin{figure}[b!]
\centering
\includegraphics[width=1.0\linewidth]{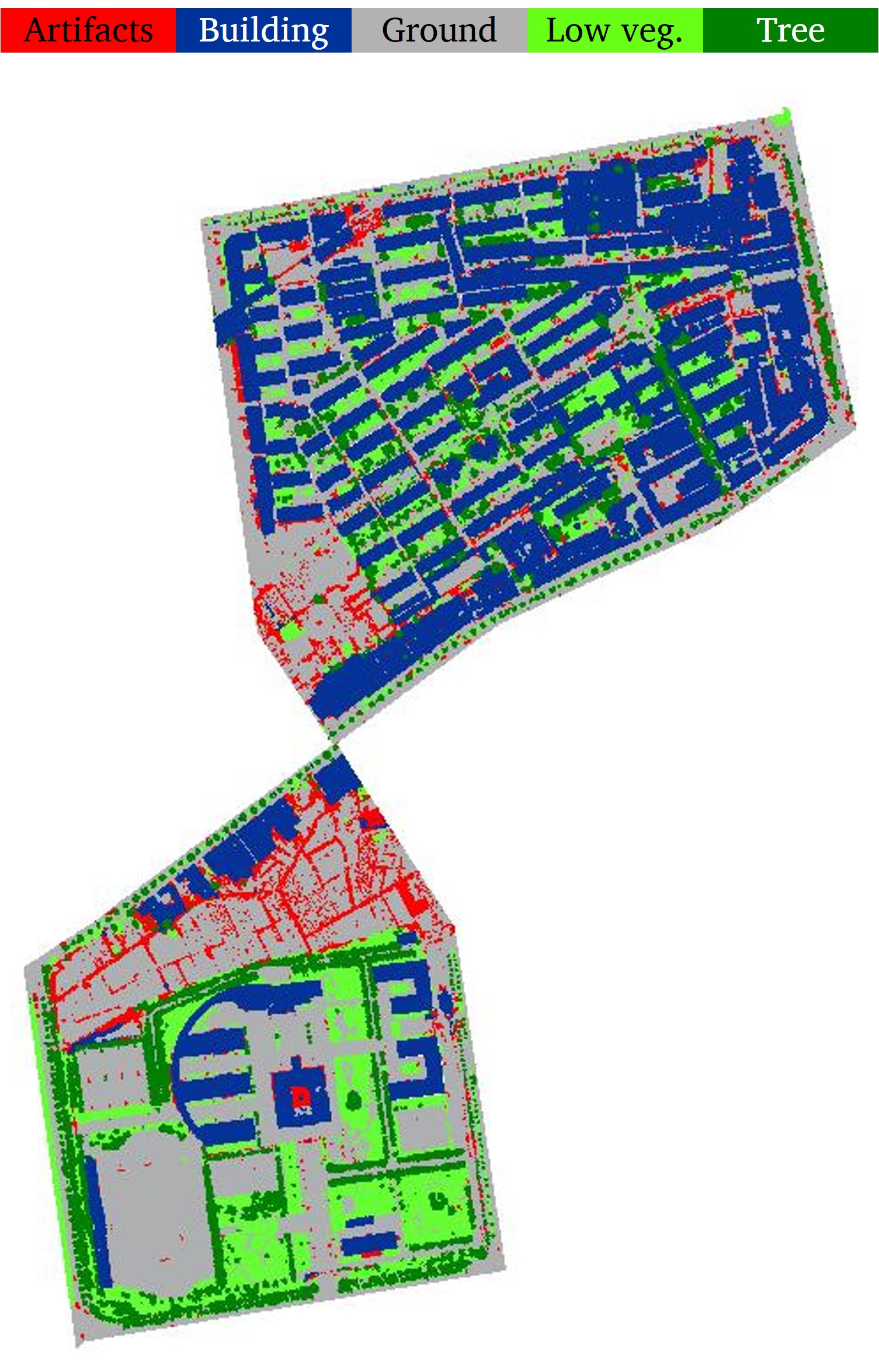}
\figcaption{The classification result at 1\textperthousand{} weak-label on LASDU dataset.}
\label{fig:lasdu_result}
\end{figure}

\begin{figure*}[t!]
\begin{center}
      \subfigure[]{
                \begin{minipage}{.3\linewidth}
                \centering
                \includegraphics[width=1.0\columnwidth]{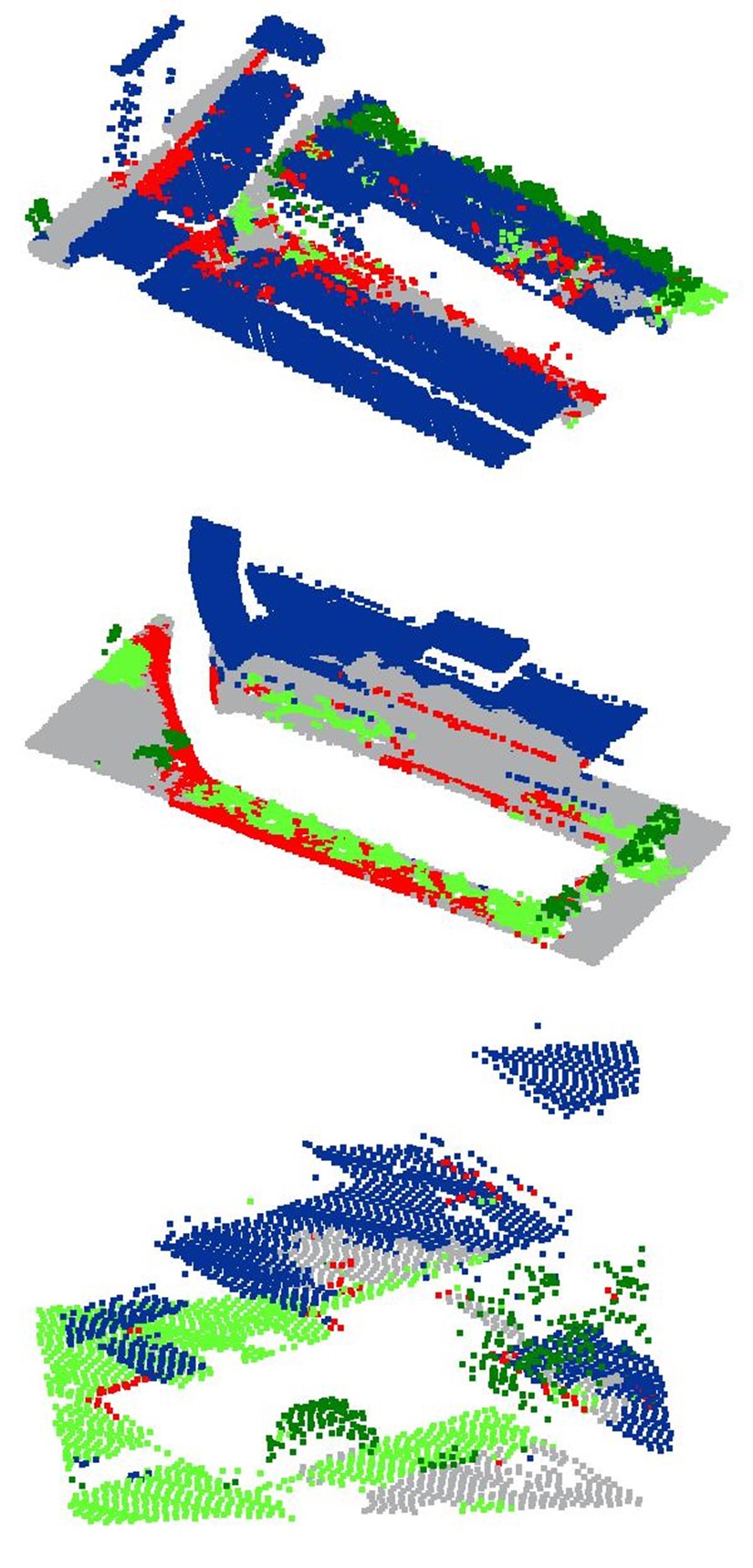}
                \end{minipage}
                }
      \subfigure[]{
                \begin{minipage}{.3\linewidth}
                \centering
                \includegraphics[width=1.0\columnwidth]{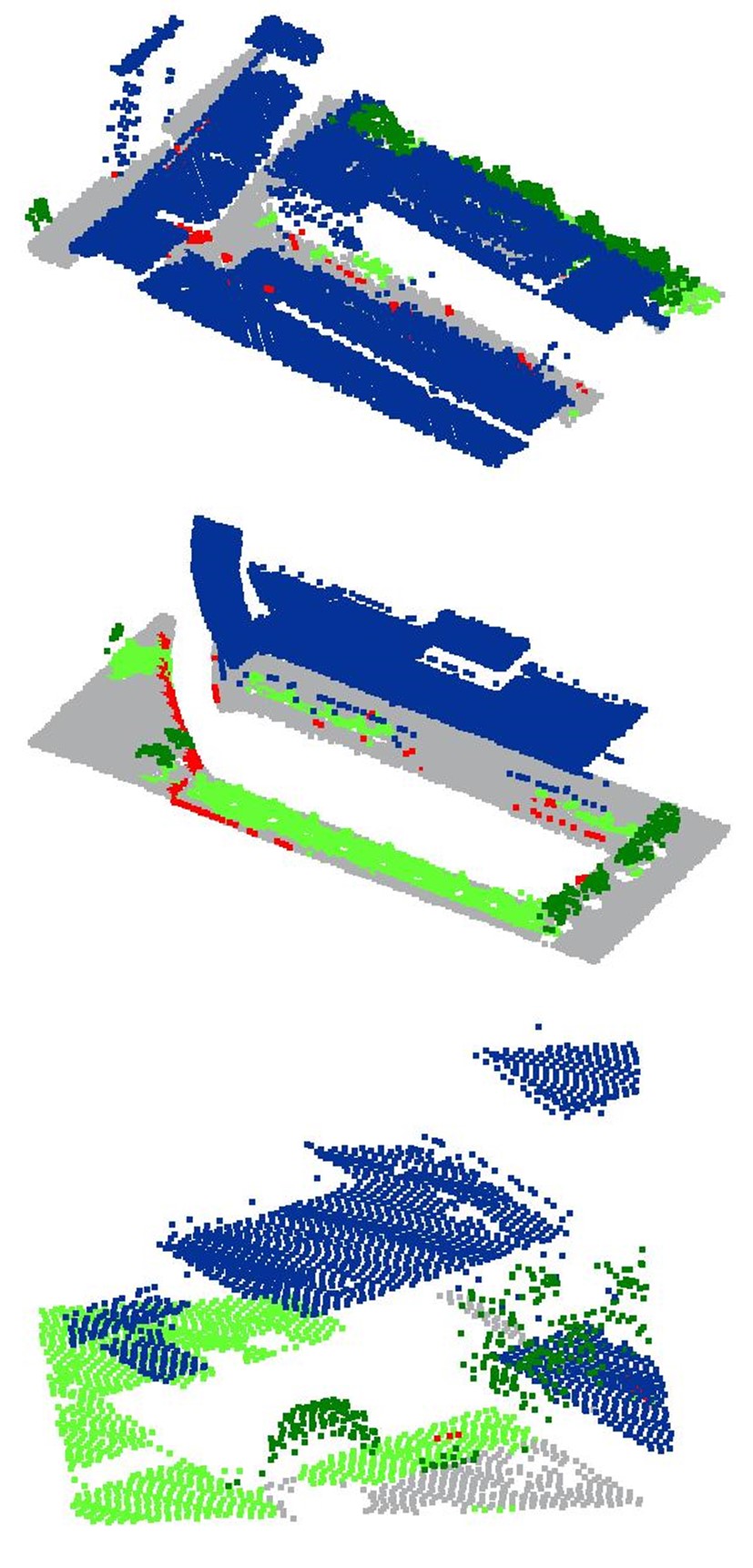}
                \end{minipage}
                } 
    \subfigure[]{
                \begin{minipage}{.3\linewidth}
                \centering
                \includegraphics[width=1.0\columnwidth]{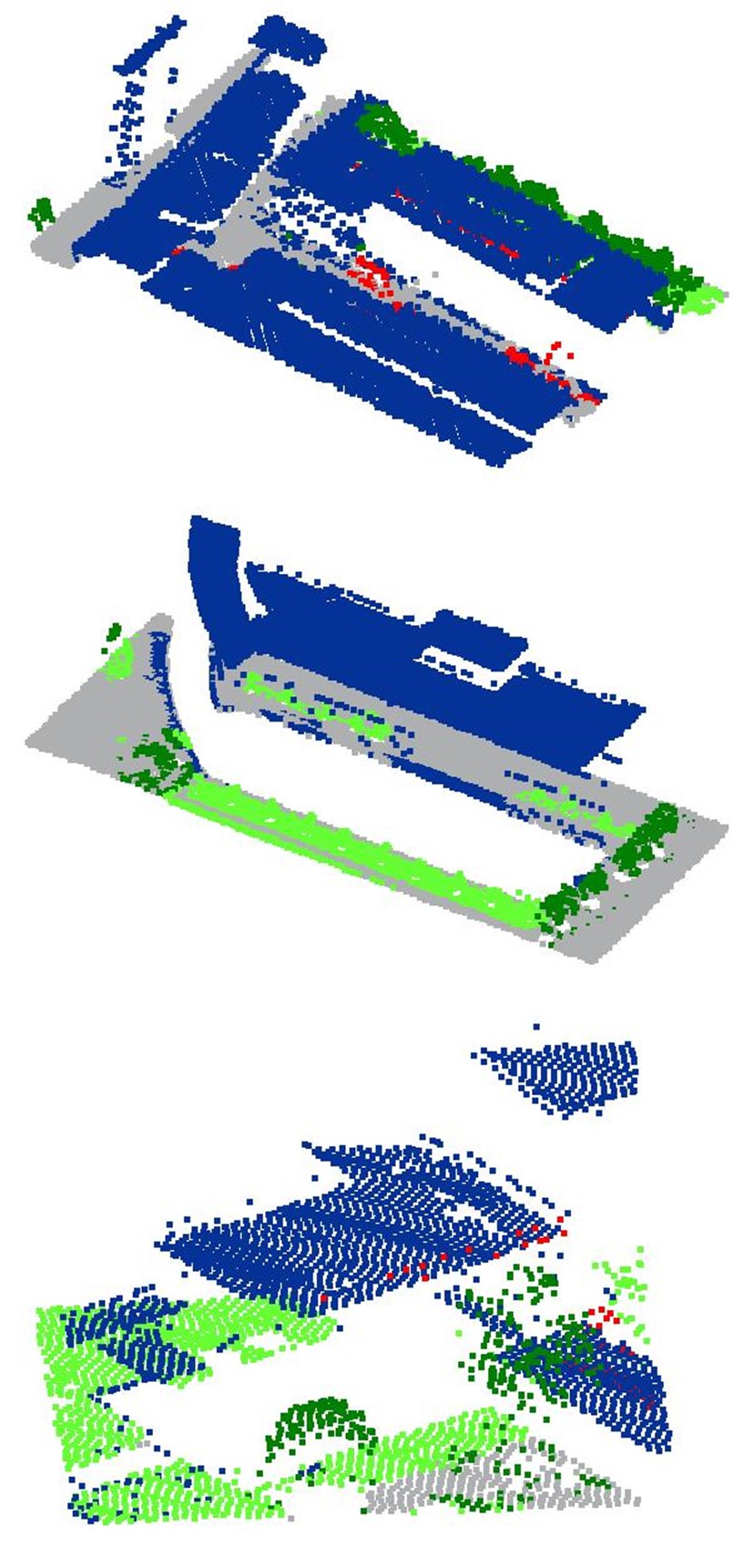}
                \end{minipage}
                } 
	\figcaption{Details of classification result on LASDU dataset. (a) and (b) represent the results of baseline and our method, and (c) is the ground truth. }
\label{fig:lasdu_local}
\end{center}
\end{figure*}

Quantitative comparison results are listed in Table~\ref{tab:isprs}.
Using 1\textperthousand{} of labels, compared with baseline, OA and F1 score of every category have considerably improved using our method. By contrast, there is only a slightly rise in OA and average F1 score for MT. This may be related to the characteristics of ISPRS dataset. As the number of points is fairly small, each point is fed into the network at a high frequency. The model converges quickly using limited weak labels, hence there is hardly difference between model parameters and their EMA value, which limits the efficacy of the consistency constraint. Xu \& Lee achieves a unsatisfactory classification performance. While one reason is the gap between different backbone networks, we argue that the inexact supervision branch cannot cope with fairly sparse weak labels, which is discussed in Section~\ref{sec:contextual}. As for methods under full supervision, we can first see that there is only a small gap for the F1 score of each category between our method and KPConv, which means our weakly supervised method can strengthen the model to extract deep features similar to those from full supervision. NANJ2 and WhuY4 achieve a higher OA, but avrage F1 score is still lower than our method. Two novel methods, GANet and GraNet, achieved more accurate classification result. Compared to them, the gap are mainly attributed to classes of fence and shrub. 

\subsubsection{LASDU dataset}
 
Following the same comparison strategy, we first present the increment by our method, presented in Fig.~\ref{fig:lasdu_line}. Compared to baseline, OA and average F1 score are considerately improved by our method. However, the growth trends of the two indicators are different. While average F1 score has improved by about 4\% under different weak label ratios, the OA increases up to 87.3\% in 5\textpertenthousand{} of weak labels, but showing only a marginal rise when increasing the number of labels. Additionally, both OA and average F1 score using 5\% of labels are still lower than that of full supervision. As there are only 5 main classes in the dataset, some of which contain indeed subcategories, the intraclass feature dissimilarity is comparatively large. Then, it is less feasible for weak labels to capture the overall information of the category under the same ratio. We still show a classification map using 1\textperthousand{} of labels, presented in Fig.\ref{fig:lasdu_result}. It can be seen that four main categories are distinguished well, whereas more errors are shown in artifacts. Despite this, our method performs much better than baseline, and Fig.~\ref{fig:lasdu_local} illustrates a detailed results in several regions. Our method rectified most of classification noises from baseline, so that the overall results exhibit high smoothness.

\begin{table*}[t!]
\tablecaption{Comparison of full and weak supervision scheme on LASDU dataset}
\label{tab:lasdu}
\centering
\footnotesize
\setlength\tabcolsep{2pt}%
\begin{tabular}{ccccccccc}
\hline
\multirow{2}{*}{Setting}                                                  & \multirow{2}{*}{Method} & \multicolumn{5}{c}{F1 Score}                   & \multirow{2}{*}{Avg. F1} & \multirow{2}{*}{OA} \\ \cline{3-7}
                                                                          &                         & Artifacts & Building & Ground & Low veg & Tree &                          &                     \\ \hline
\multirow{3}{*}{Full Sup.}                                               & RandLA-Net              & 46.2      & \textbf{96.4}     & 92.0   & 70.2    & \textbf{87.3} & 78.4                     & 88.5                \\
                                                                          & KPConv                  & \textbf{46.9}      & 96.3     & \textbf{92.3}  & \textbf{73.4}    & 86.9 & \textbf{79.1}                     & \textbf{88.9}                \\
                                                                          & GrabNet                 & 42.4      & 95.8     & 89.9   & 64.7    & 86.1 & 75.8                     & 86.2                \\ \hline
\multirow{4}{*}{\begin{tabular}[c]{@{}c@{}}Weak Sup.\\ (1\textperthousand{})\end{tabular}} & Xu \& Lee                      & 15.9      & 90.8     & 88.5   & 61.5    & 72.3 & 65.8                     & 82.4                \\
                                                                          & MT                      & \textbf{40.7}      & 94.2     & 88.3   & 72.3    & 82.0 & 75.5                     & 84.4                \\
                                                                          & Baseline                & 35.3      & 92.7     & 86.1   & 67.1    & 84.5 & 73.1                     & 81.4                \\
                                                                          & Ours                    & 40.4      & \textbf{95.0}     & \textbf{90.9}   & \textbf{72.4}    & \textbf{88.3} & \textbf{77.4}                     & \textbf{87.6}                \\ \hline
\end{tabular}
\end{table*}

\begin{figure*}[t!]
\centering
\includegraphics[width=0.85\linewidth]{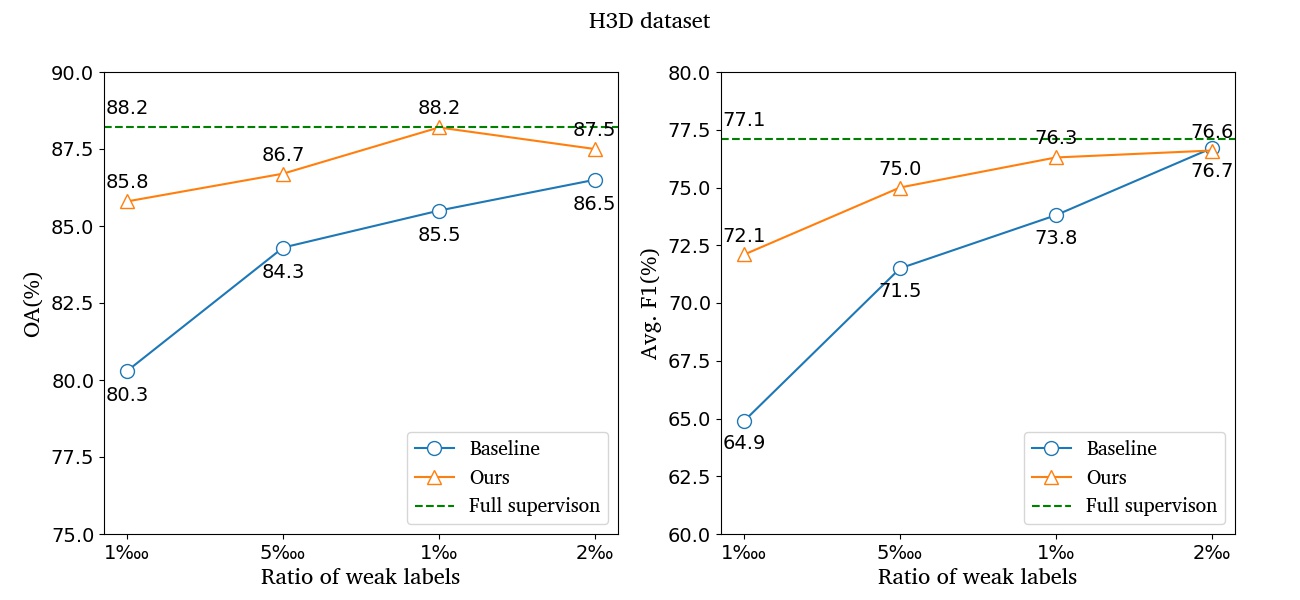}
\figcaption{The impact of the ratio of labeled points on OA and Avg. F1 on H3D dataset.}
\label{fig:h3d_line}
\end{figure*}

\begin{figure}[t]
\includegraphics[width=1.0\linewidth]{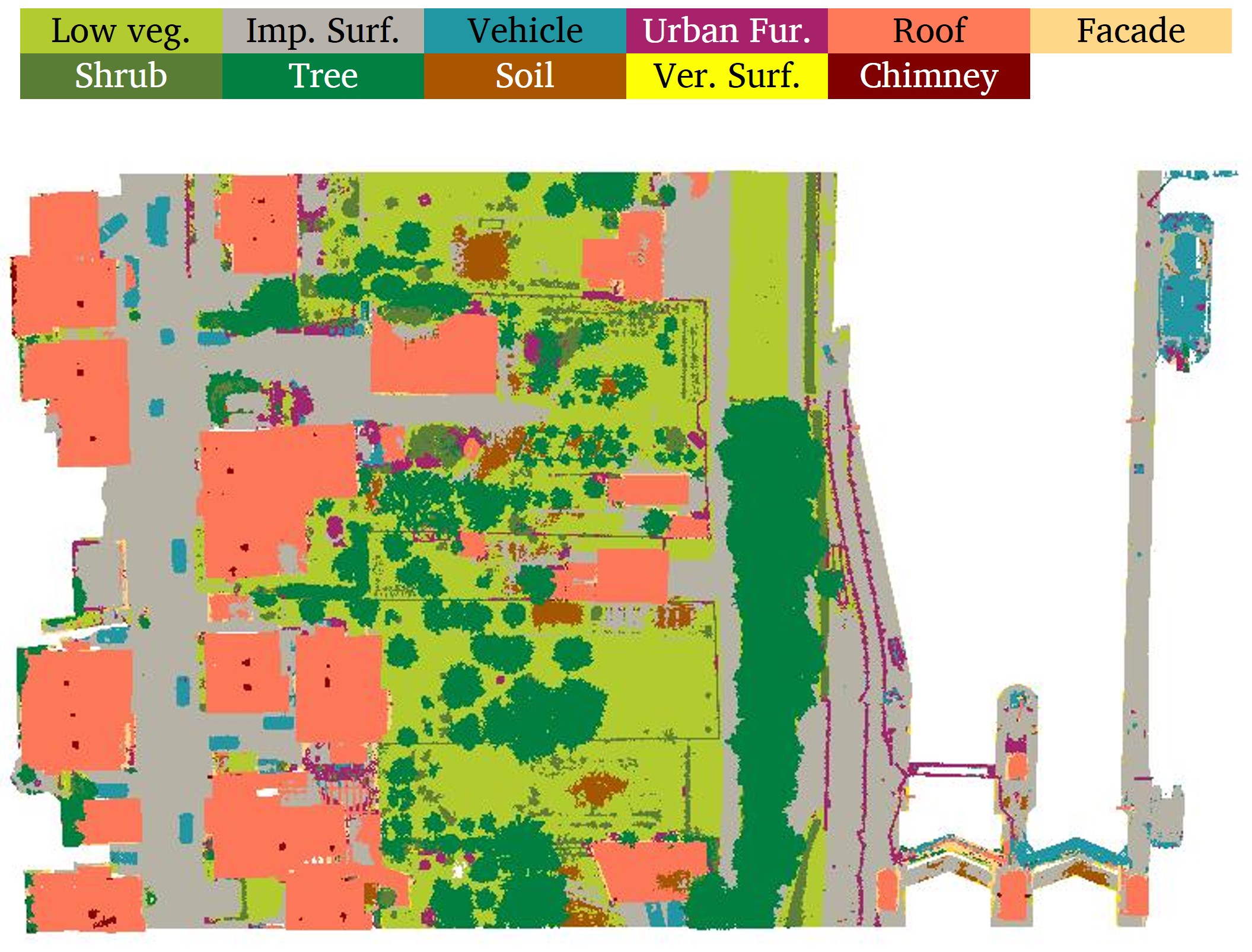}
\figcaption{The classification result at 1\textperthousand{} of weak-label on H3D dataset.}
\label{fig:h3d_result}
\end{figure}

Several other methods are compared here, and quantitative results are listed in Table~\ref{tab:lasdu}. Similarly, we analyze weakly supervised learning methods first. Using 1\textperthousand{} of labels, our method achieves a considerable rise in OA and F1 score of every category compared with baseline. MT acquires a much better result than baseline on LASDU dataset, but worse than our method, and the gap mainly exists in the classes of tree and ground. By contrast, the performance of Xu \& Lee is still poor. Comparing with results under full supervision, we can see that our method outperforms GrabNet, slightly below RandLA-Net. Our backbone netowrk, KPConv, acquires the highest result. The classification error of artifacts from our method is the main issue leading to the performance gap.

\begin{figure}[t!]
\begin{center}
      \subfigure[]{
                \begin{minipage}{.3\linewidth}
                \centering
                \includegraphics[width=1.1\columnwidth]{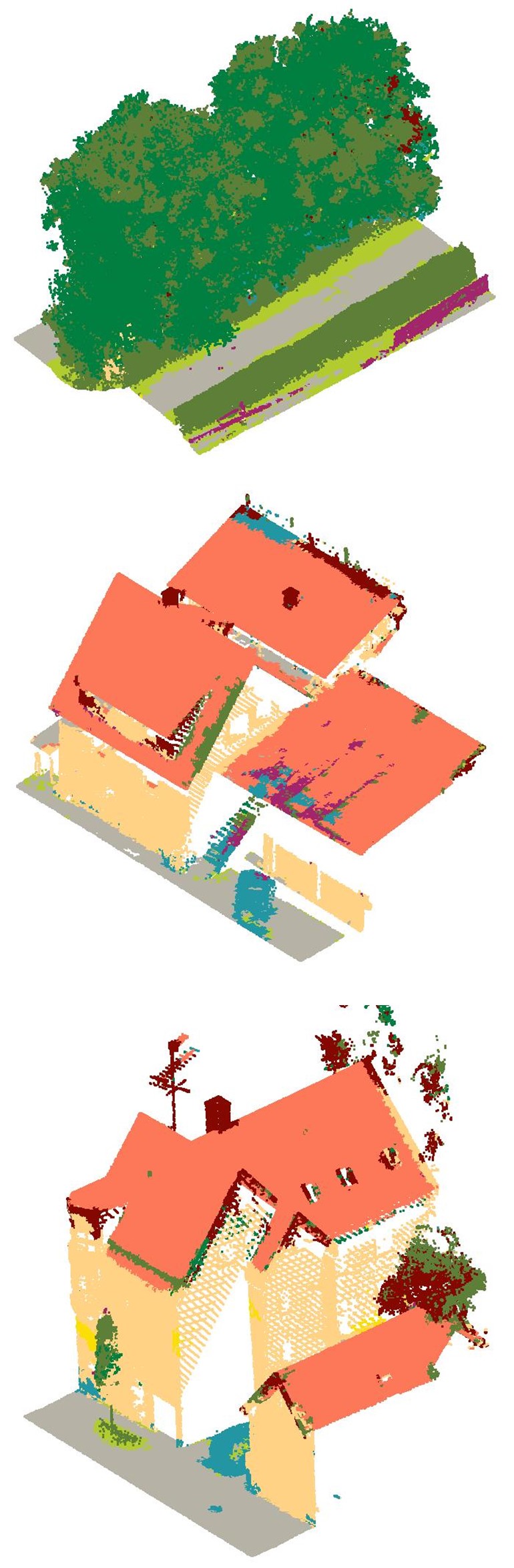}
                \end{minipage}
                }
      \subfigure[]{
                \begin{minipage}{.3\linewidth}
                \centering
                \includegraphics[width=1.1\columnwidth]{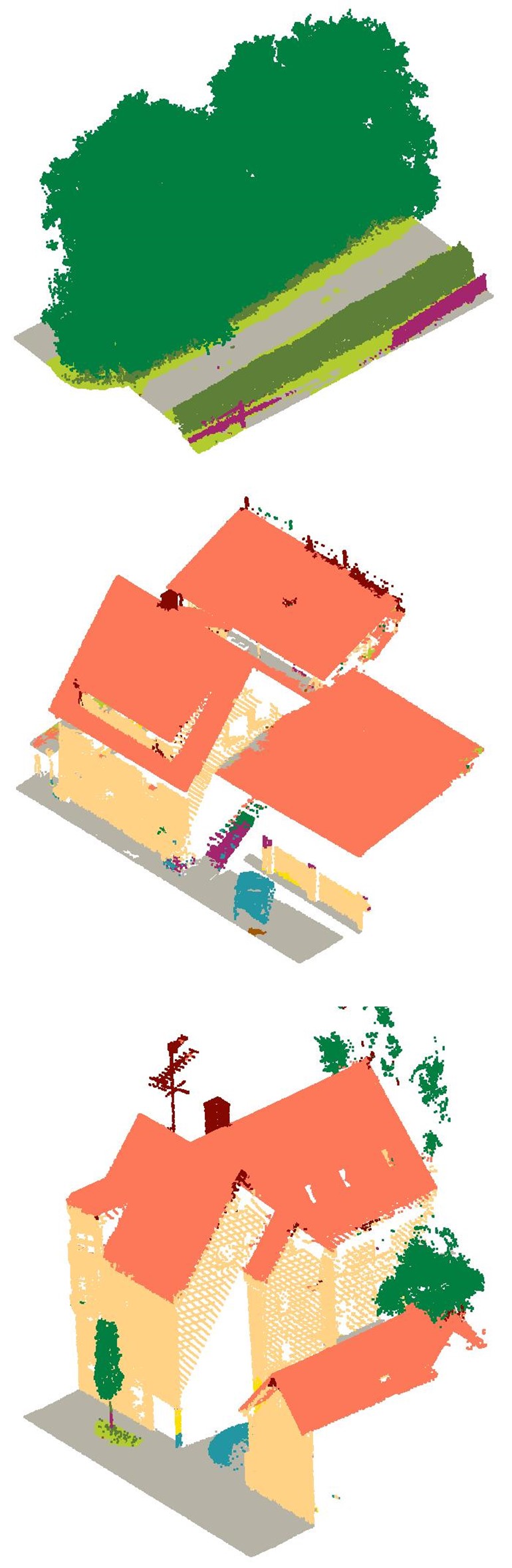}
                \end{minipage}
                } 
    \subfigure[]{
                \begin{minipage}{.3\linewidth}
                \centering
                \includegraphics[width=1.1\columnwidth]{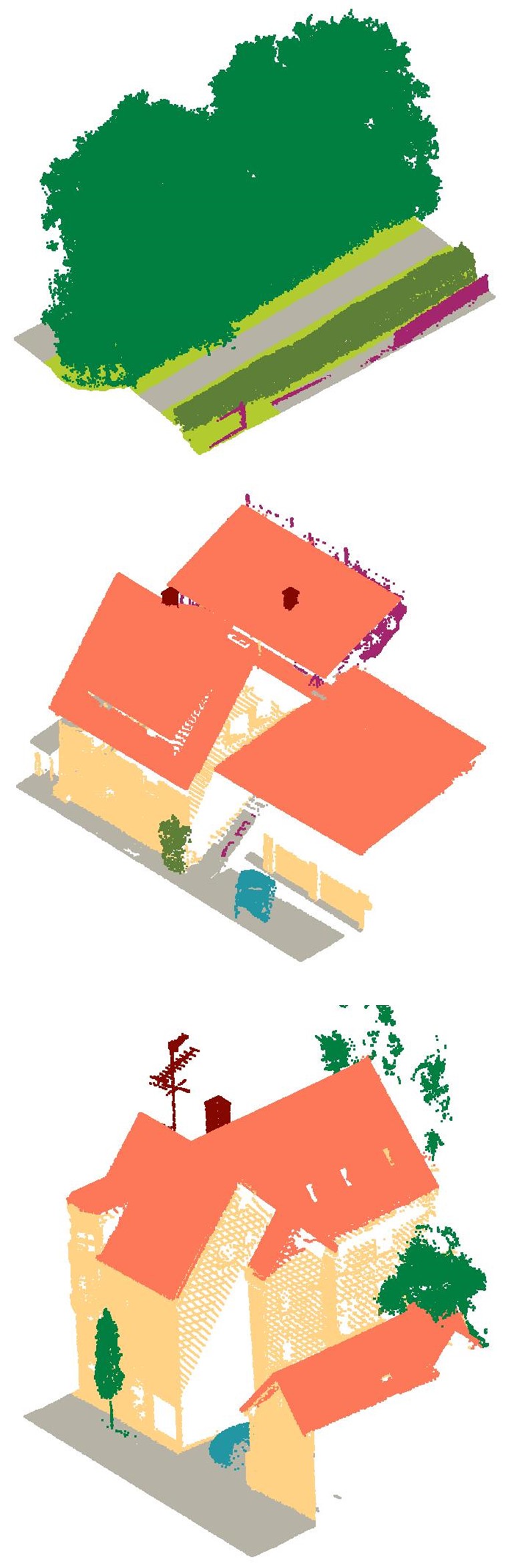}
                \end{minipage}
                } 
	\figcaption{Details of classification result on H3D dataset. (a) and (b) represent the results of baseline and our method, and (c) is the ground truth. }
\label{fig:h3d_local}
\end{center}
\end{figure}

\subsubsection{H3D dataset}

The performance improvement incurred by our method is presented in Fig.~\ref{fig:h3d_line}, which shows that there is a rise in OA and average F1 score for most of weak-label settings. Our method attains the same OA as that of full supervision, and marginally lower average F1 score using 1\textperthousand{} of weak labels. However, slightly poorer results can be obtained using 2\textperthousand{} of labels. The reason may be that 1\textperthousand{} of weak labels is adequate for saturating the performance limit of the backbone network, leading results to have a small range of deviation. Another noteworthy finding is that the average F1 score of baseline at 2\textperthousand{} of weak labels almost reaches the level of full supervision. Owing to fact that the density of H3D dataset is greatly higher than other two datasets, a large number of weak labels exist. Thus, we believe that for H3D dataset a large number of redundant annotations exist as well. In Fig.~\ref{fig:h3d_result}, we provide the classification result using 1\textperthousand{} of weak labels. Due to the high point density and the distinct geometric structure of objects, the classification result also maintain good boundary information. We further make a detailed comparison between baseline and our method in Fig.~\ref{fig:h3d_local}, where the first row shows a few trees. Due to the similar feature information under weak labels, some points are misclassified to shrub, which are corrected by our method. Meanwhile, classification errors on roof are also rectified as shown in the other two rows. 

Quantitative comparison results are listed in Table~\ref{tab:h3d}, where we analyze the results using 1\textperthousand{} and 1\textpertenthousand{} of labels. Under two weak-label settings, our method achieves a considerable rise in OA and F1 score for most of categories compared with baseline. The performance gap between our method and MT reduces on this dataset. Since MT updates model parameters per step, it may help maintain high robustness in case of large number of training samples, thus achieving better result. Xu \& Lee performs poorly at some classes, such as chimney and shrub. Using 1\textperthousand{} of labels, our method obtains the comparable result with RandlA-Net, while the result of shrub and vertical surface poses main problem leading to lower average F1 score compared to KPConv.

\begin{table*}[t]
\tablecaption{Comparison of full and weak supervisions on H3D dataset}
\label{tab:h3d}
\centering
\footnotesize
\setlength\tabcolsep{2pt}%
\begin{tabular}{ccccccccccccccc}
\hline
\multirow{2}{*}{Setting}                                                         & \multirow{2}{*}{Method} & \multicolumn{11}{c}{F1 Score}                                                                             & \multirow{2}{*}{Avg. F1} & \multirow{2}{*}{OA} \\ \cline{3-13}
                                                                                 &                         & {\begin{tabular}[c]{@{}c@{}}Low \\ veg.\end{tabular}} & {\begin{tabular}[c]{@{}c@{}} Imp.\\ Surf.\end{tabular}} & Vehicle & {\begin{tabular}[c]{@{}c@{}} Urban\\ Fur. \end{tabular}}& Roof & Facade & Shrub & Tree & Soil & {\begin{tabular}[c]{@{}c@{}}Ver.\\Surf. \end{tabular}} & Chimney &                          &                     \\ \hline
\multirow{2}{*}{Full Sup.}                                                   & RandLA-Net              & \textbf{89.0}     & \textbf{90.5}       & 47.8    & 63.1       & \textbf{96.8} & 79.0   & 62.2  & 95.3 & \textbf{45.5} & 74.5       & \textbf{87.2}    & 75.5                     & 88.1                \\
                                                                                 & KPConv                  & 88.3     & 90.0       & \textbf{66.2}    & \textbf{63.8}       & 96.5 & \textbf{81.8}   & \textbf{68.6}  & \textbf{95.7} & 32.5 & \textbf{81.7}       & 83.3    & \textbf{77.1}                     & \textbf{88.2}                \\ \hline
\multirow{4}{*}{\begin{tabular}[c]{@{}c@{}}Weak Sup.    \\ (1\textperthousand{})\end{tabular}} & Xu \& Lee                     & 83.1     & 75.3       & 42.9    & 40.3       & 91.4 & 60.2   & 23.8  & 90.0 & 34.0 & 58.1       & 0.1     & 55.4                     & 81.2                \\
                                                                                 & MT                      & 87.4     & \textbf{89.5}       & \textbf{75.2}    & 62.8       & 96.2 & 80.2   & \textbf{60.4}  & \textbf{95.6} & \textbf{43.8} & \textbf{71.7}       & 70.9    & 75.8                     & 87.3                \\
                                                                                 & Baseline                & 86.1     & 88.3       & 71.2    & 63.4       & 93.4 & \textbf{81.1}   & 59.4  & 94.7 & 38.5 & 70.3       & 65.7    & 73.8                     & 85.5                \\
                                                                                 & Ours                    & \textbf{89.4}     & 89.2       & 73.9    & \textbf{63.6}       & \textbf{96.4} & 79.9   & 60.1  & 95.1 & 41.9 & 69.3       & \textbf{80.7}    & \textbf{76.3}                     & \textbf{88.2}                \\ \hline
\multirow{3}{*}{\begin{tabular}[c]{@{}c@{}}Weak Sup. \\     (1\textpertenthousand{})\end{tabular}}  &MT                      & 85.0     & 85.7       & 63.1    & 51.0       & 93.9 & 76.0   & \textbf{55.4}  & \textbf{94.0} & 35.8 & 65.5       & 50.8    & 68.8                     & 84.0                \\
                                                                                 & Baseline                & 83.5     & 84.7       & 58.0    & 46.5       & 92.3 & 76.1   & 54.1  & 92.6 & \textbf{39.7} & 65.7       & 46.7    & 64.9                     & 80.3                \\
                                                                                 & Ours                    & \textbf{87.2}     & \textbf{86.9}       & \textbf{67.1}    & \textbf{53.4}       & \textbf{94.8} & \textbf{80.2}   & 50.5  & 93.9 & 33.6 & \textbf{73.3}       & \textbf{71.9}    & \textbf{72.1}                     & \textbf{85.8}                \\ \hline
\end{tabular}
\end{table*}

\subsection{Ablation study}
The effectiveness of each module of proposed method is discussed in this section, and a comparison of the experimental results is presented in Table~\ref{tab:ablation}. The experiments are conducted using 1\textperthousand{} of labels for ISPRS and LASDU dataset, and 1\textpertenthousand{} for H3D dataset. We first analyze the improvement by each of individual components. From the table, it shows considerable increases for all three modules on all datasets. On ISPRS dataset, both ER and OSPL increase OA by 5\%+ and average F1 score by 10\%+ compared with the baseline. By contrast, the growth rate in EPC is smaller, and the reason is similar to MT. On LASDU dataset, the increment for OA and average F1 score in every module equals to about 4\% and 5\%. On H3D dataset, we can see the OA of EPC and OSPL have obviously improved compared to that of ER. Comparing the performance of each individual modules using different datasets, it indicates that as the number of weak labels in the dataset increases, the effect of ER becomes gradually insignificant, whereas that of EPC becomes increasingly obvious and that of OSPL remains stable. It could be caused by different characteristics of each module. The increased number of weak labels will alleivate the underfitted training model. Then, usable information contained in unlabeled samples decreases, which leads to the degradation effect of ER. By contrast, though OSPL also directly utilizes unlabeled samples to generate pseudo-labels, the accuracy of pseudo-labels grows as the number of weak labels is increased, thus allowing OSPL to maintain a steady rise in evaluation indices. As for EPC, large amount of points gives rise to great diversity in mini-batches, which contributes more robust ensemble value to consistency constraint. Despite the effect of each individual module, it still shows a further rise in evaluation metrics when combining these modules. Due to the similarity, we combine ER and OSPL to compare with our method. From the table, it shows that the combined version outperforms individual components on all three datasets, and our method achieves the best result in all settings, which demonstrates the effectiveness of our method. 

\begin{table}[h]
\tablecaption{Effectiveness of Entorpy Regularization (ER), ensemble prediction constraint (EPC) and online soft pseudo-labeling (OSPL) on three datasets}
\label{tab:ablation}
\centering
\footnotesize
\setlength\tabcolsep{2pt}%
\begin{tabular}{ccccccccccc}
\hline
\multicolumn{5}{c}{Module}                                 & \multicolumn{2}{c}{ISPRS} & \multicolumn{2}{c}{LASDU} & \multicolumn{2}{c}{H3D} \\ \hline
ER    & \multicolumn{2}{c}{EPC} & \multicolumn{2}{c}{OSPL} & Avg. F1       & OA        & Avg. F1       & OA        & Avg. F1      & OA       \\ \hline
      & \multicolumn{2}{c}{}    & \multicolumn{2}{c}{}     & 57.2          & 76.1      & 73.1          & 81.4      & 64.9         
      & 80.3     \\
\checkmark     & \multicolumn{2}{c}{}    & \multicolumn{2}{c}{}     & 67.5          & 81.2      & 77.0          & 85.7      & 68.7         & 81.5     \\
      & \multicolumn{2}{c}{\checkmark}   & \multicolumn{2}{c}{}     & 61.1          & 78.9      & 76.7          & 87.2      & 68.1         & 83.6     \\
      & \multicolumn{2}{c}{}    & \multicolumn{2}{c}{\checkmark}    & 68.6          & 81.5      & 77.0          & 87.3      & 68.5         & 83.0     \\
\checkmark     & \multicolumn{2}{c}{}   & \multicolumn{2}{c}{\checkmark}     & 68.7          & 82.6      & \textbf{77.4}          & 87.5      & 69.2         & 83.8     \\
\checkmark     & \multicolumn{2}{c}{\checkmark}   & \multicolumn{2}{c}{\checkmark}    & \textbf{70.0}          & \textbf{83.0}      & \textbf{77.4}          & \textbf{87.6}      & \textbf{72.1}         & \textbf{85.8}     \\ \hline
\end{tabular}
\end{table}

\begin{table}[b]
\tablecaption{A comparison of training time on different datasets}
\label{tab:time}
\centering
\footnotesize
\setlength\tabcolsep{2pt}%
\begin{tabular}{cccc}
\hline
\multirow{2}{*}{Method} & \multicolumn{3}{c}{Running time (h)} \\ \cline{2-4} 
                         & ISPRS       & LASDU       & H3D      \\ \hline
Baseline                 & 1.9         & 4.4         & 5.6      \\
MT                       & 2.6(+37\%)         & 6.1(+39\%)         & 8.1(+45\%)      \\
Ours                     & 2.3(+21\%)         & 5.3(+20\%)         & 7.3(+30\%)      \\ \hline
\end{tabular}
\end{table}

\subsection{Complexity and runtime analysis}
We consider baseline, MT and our method, and the comparison is presented in Table~\ref{tab:time}. All three methods use KPConv as the backbone network. Our method makes use of output predictions during training, incurring no increase in model size. As for MT, though it needs to create shadow variables to save EMA value of model parameters, those variables do not participate in loss calculation and backpropagation. Thus, both our method and MT have the same model complexity compared to baseline, and the comparison is not listed. The running time refers to the whole training process, including a period of validation after each epoch. Under the same implementation condition, our method takes up 21\% and 20\% more running time on ISPRS and LASDU dataset compared to baseline, respectively, and for the case of H3D dataset it is increased by up to 30\%. The reason is that EMA value of predictions at each step needs to be calculated and saved. By contrast, MT is obviously more demanding in terms of  running time because not only model parameters need to be updated at each training step, but also two forward propagation are required to generate contrastive prediction pairs. Thus, our method shows better running efficiency compared to MT. It should be noticed that only training time is discussed here. Owing to fact that there is no modification in network structure design, the testing process of all three methods remains the same.

\subsection{Comparison to exploiting contextual information}\label{sec:contextual}
Contextual information describes overall class distribution in a scene. In this study, it refers specifically to existed scene-level labels in one training sample. ~\citet{liu2020fg} proposed a contextual loss for point cloud scene understanding. In~\citet{xu2020weakly}, it was utilized as inexact supervision using sparse weak labels. The rational behind that branch was that an object category should be contained even if only one weak label of the category is present in the sample. While experimental results showed the effectiveness using 10\% of total labels in that work, we argue that it could produce incorrect information using fairly limited weak labels, which undermines the model training. When the number of weak labels continues to decrease, one situation occurs that no annotation for some existent categories in the sample can be acquired. The incorrect contextual information will lead to degraded effect because the loss function rewards the probability of existed classes while penalize that of nonexistent ones. A further experiment is presented in Table~\ref{tab:context}, where the effect of contextual loss is compared using datasets in this study. From the table, we can see a huge drop in accuracy on ISPRS dataset and a decline for other two. The characteristics of the dataset is the main reason, since low density, small number of points and relatively large number of categories of ISPRS dataset makes it difficult for weak labels to reflect complete scene-level label information.

\begin{table}[h]
\tablecaption{Comparison with results using contextual information (CI) using 1\textperthousand{} of labels on different datasets.}
\label{tab:context}
\centering
\footnotesize
\setlength\tabcolsep{2pt}%
\begin{tabular}{ccccccc}
\hline
\multirow{2}{*}{Method}                                                & \multicolumn{2}{c}{ISPRS} & \multicolumn{2}{c}{LASDU} & \multicolumn{2}{c}{H3D} \\ \cline{2-7} 
                                                                       & Avg. F1       & OA        & Avg. F1       & OA        & Avg. F1      & OA       \\ \hline
Ours                                                                   & \textbf{70.0}          & \textbf{83.0}      & \textbf{77.4}          & \textbf{87.6}      & 76.3         & \textbf{88.2}     \\
Ours+CI & 58.0          & 81.7      & 76.7          & 87.2      & \textbf{77.2}         & 87.3     \\ \hline
\end{tabular}
\end{table}

\subsection{Limitation}
Our method focuses on utilizing potential information in unlabeled data and proposes three prediction-constraint strategies to improve classification accuracy under weak supervision. Though it is a lightweight framework, which can be easily integrated to diverse networks, we believe that a pre-designed network architecture targeted for weak supervision could further promote the development of this field. An experiment on ISPRS dataset is illustrated in Table~\ref{tab:backbone}, in which two proven network, KPConv and RandLA-Net, are considered for comparing their performance under 1\textperthousand{} of labels. While there exists only a small performance gap between two fully supervised models, a large difference is observed under weak supervision. Thus, we argue that different networks are more sensitive to respond to weakly supervised point cloud classification tasks, especially for those fairly sparse labels. Though the attention mechanism, which successfully applied in many studies, is considered in our experiments, no direct solution to changing the network architecture is presented. We integrate DualAttention~\citep{fu2019dual} module into KPConv with an intention to enhance the ability of feature representation. Considering the limitation of computational resource, it is only added to the last layer of the encoder network, while experimental results do not show a substantial improvement from Table~\ref{tab:attention}. Thus, it seems that it does not bring much benefit when merely adding a general attention module, and it becomes our future work to propose a network closely compatible to weak supervision.

\begin{table}[h]
\tablecaption{Classification results using different backbone networks.}
\label{tab:backbone}
\centering
\footnotesize
\setlength\tabcolsep{2pt}%
\begin{tabular}{ccccc}
\hline
Method                      & \multicolumn{2}{c}{Setting}           & Avg. F1 & OA   \\ \hline
\multirow{3}{*}{KPConv}     & \multicolumn{2}{c}{Full Sup.}         & 71.2    & 83.8 \\ \cline{2-5} 
                            & \multirow{2}{*}{Weak Sup.} & Baseline & 57.2    & 76.1 \\
                            &                            & Ours     & 70.0    & 83.0 \\ \hline
\multirow{3}{*}{RandLA-Net} & \multicolumn{2}{c}{Full Sup.}         & 70.1    & 83.1 \\ \cline{2-5} 
                            & \multirow{2}{*}{Weak Sup.} & Baseline & 52.3    & 70.6 \\
                            &                            & Ours     & 64.1    & 78.5 \\ \hline
\end{tabular}
\end{table}

\begin{table}[h]
\tablecaption{Comparison with results integrating an attention module, DualAttention (DA)}
\label{tab:attention}
\centering
\footnotesize
\setlength\tabcolsep{2pt}%
\begin{tabular}{ccccccc}
\hline
\multirow{2}{*}{Method} & \multicolumn{2}{c}{ISPRS} & \multicolumn{2}{c}{LASDU} & \multicolumn{2}{c}{H3D} \\ \cline{2-7} 
                        & Avg. F1       & OA        & Avg. F1       & OA        & Avg. F1      & OA       \\ \hline
Ours                    & \textbf{70.0}          & 83.0      & \textbf{77.4}          & \textbf{87.6}      & 76.3         & 88.2     \\
Ours + DA    & 69.7          & \textbf{83.1}      & 77.1          & 87.3      & \textbf{76.5}         & \textbf{88.3}     \\ \hline
\end{tabular}
\end{table}

\section{Conclusion}\label{sec:conclusion}

In this study, we investigate semantic segmentation of ALS point clouds using sparse annotations and propose an efficient weakly supervised framework, which is compatible with current point cloud classification networks. An entropy regularization module is introduced to reduce class overlap in predictions and improve the classification confidence. An ensemble prediction constraint, where a consistency loss is calculated between prediction confidence at current training step and its ensemble value, is proposed to strength the robustness of predictions. Additionally, an online soft pseudo-labeling module is developed to take advantage of output predictions of training samples and supply extra supervisory sources. We perform comprehensive experiments to evaluate our method using three benchmark ALS datasets. Our method significantly improves OA and average F1 score compared with baseline and achieves comparable results against the full supervision competitors using only 1\textperthousand{} of labels. Experimental results demonstrate that our method can help to largely reduce the workload of data annotation as only sparse labels randomly generated across the scene are needed to attain promising results.

In the future, we would like to solve the limitation of our method as discussed before. A well-designed network architecture closely compatible to weak supervision could better exploit contextual relations between sparse weak labels. In addition, random selection may not the best way to initialize weak labels. Unsupervised pre-training could assist with prior knowledge and enable the model to achieve better performance using the same number of annotations.


\section*{Acknowledgements}
The ISPRS dataset was provided by the German Society for Photogrammetry, Remote Sensing, and Geoinformation (DGPF). The LASDU dataset was provided by College of Surveying and Geo-informatics, Tongji University and Photogrammetry and Remote Sensing, Technical University of Munich. The H3D dataset was provided by Institute for Photogrammetry, University of Stuttgart.










\bibliographystyle{elsarticle-harv}
\bibliography{mybibfile}

\end{document}